\PassOptionsToPackage{unicode}{hyperref}
\PassOptionsToPackage{hyphens}{url}
\PassOptionsToPackage{dvipsnames,svgnames,x11names}{xcolor}
\documentclass[
  11pt,
  a4paper,
]{article}
\usepackage{amsmath,amssymb}
\usepackage{iftex}
\ifPDFTeX
  \usepackage[T1]{fontenc}
  \usepackage[utf8]{inputenc}
  \DeclareUnicodeCharacter{2212}{\ensuremath{-}}
  \usepackage{textcomp} % provide euro and other symbols
\else % if luatex or xetex
  \usepackage{unicode-math} % this also loads fontspec
  \defaultfontfeatures{Scale=MatchLowercase}
  \defaultfontfeatures[\rmfamily]{Ligatures=TeX,Scale=1}
\fi
\usepackage{lmodern}
\IfFileExists{upquote.sty}{\usepackage{upquote}}{}
\IfFileExists{microtype.sty}{% use microtype if available
  \usepackage[]{microtype}
  \UseMicrotypeSet[protrusion]{basicmath} % disable protrusion for tt fonts
}{}
\makeatletter
\@ifundefined{KOMAClassName}{% if non-KOMA class
  \IfFileExists{parskip.sty}{%
    \usepackage{parskip}
  }{% else
    \setlength{\parindent}{0pt}
    \setlength{\parskip}{6pt plus 2pt minus 1pt}}
}{% if KOMA class
  \KOMAoptions{parskip=half}}
\makeatother
\usepackage{xcolor}
\usepackage[margin=20mm]{geometry}
\usepackage{color}
\usepackage{fancyvrb}

\DefineVerbatimEnvironment{Highlighting}{Verbatim}{commandchars=\\\{\}}
\usepackage{framed}
\definecolor{shadecolor}{RGB}{248,248,248}
\newenvironment{Shaded}{\begin{snugshade}}{\end{snugshade}}

\newcommand{\AttributeTok}[1]{\textcolor[rgb]{0.13,0.29,0.53}{#1}}

\newcommand{\BuiltInTok}[1]{#1}

\newcommand{\CommentTok}[1]{\textcolor[rgb]{0.56,0.35,0.01}{\textit{#1}}}

\newcommand{\ControlFlowTok}[1]{\textcolor[rgb]{0.13,0.29,0.53}{\textbf{#1}}}
\newcommand{\DataTypeTok}[1]{\textcolor[rgb]{0.13,0.29,0.53}{#1}}
\newcommand{\DecValTok}[1]{\textcolor[rgb]{0.00,0.00,0.81}{#1}}

\newcommand{\ExtensionTok}[1]{#1}
\newcommand{\FloatTok}[1]{\textcolor[rgb]{0.00,0.00,0.81}{#1}}

\newcommand{\KeywordTok}[1]{\textcolor[rgb]{0.13,0.29,0.53}{\textbf{#1}}}
\newcommand{\NormalTok}[1]{#1}
\newcommand{\OperatorTok}[1]{\textcolor[rgb]{0.81,0.36,0.00}{\textbf{#1}}}

\newcommand{\StringTok}[1]{\textcolor[rgb]{0.31,0.60,0.02}{#1}}
\newcommand{\VariableTok}[1]{\textcolor[rgb]{0.00,0.00,0.00}{#1}}

\usepackage{longtable,booktabs,array}
\usepackage{calc} % for calculating minipage widths
\usepackage{etoolbox}
\makeatletter
\patchcmd\longtable{\par}{\if@noskipsec\mbox{}\fi\par}{}{}
\makeatother
\IfFileExists{footnotehyper.sty}{\usepackage{footnotehyper}}{\usepackage{footnote}}
\makesavenoteenv{longtable}
\usepackage{graphicx}
\makeatletter
\def\maxwidth{\ifdim\Gin@nat@width>\linewidth\linewidth\else\Gin@nat@width\fi}
\def\maxheight{\ifdim\Gin@nat@height>\textheight\textheight\else\Gin@nat@height\fi}
\makeatother
\setkeys{Gin}{width=\maxwidth,height=\maxheight,keepaspectratio}
\makeatletter
\def\fps@figure{htbp}
\makeatother
\newlength{\cslhangindent}
\newlength{\csllabelwidth}
\newlength{\cslentryspacingunit} % times entry-spacing
 {% don't indent paragraphs
  \setlength{\parindent}{0pt}
  \ifodd #1
  \let\oldpar\par
  \def\par{\hangindent=\cslhangindent\oldpar}
  \fi
  \setlength{\parskip}{#2\cslentryspacingunit}
 }%
 {}
\usepackage{calc}

\ifLuaTeX
\usepackage[bidi=basic]{babel}
\else
\usepackage[bidi=default]{babel}
\fi
\babelprovide[main,import]{american}
\def\languageshorthands#1{}
\usepackage{cite}
\usepackage{microtype}
\usepackage{titlesec}
\usepackage{caption}
\usepackage{placeins}
\usepackage{needspace}
\usepackage{xurl}
\usepackage{etoolbox}
\definecolor{ink}{HTML}{153C4A}
\definecolor{muted}{HTML}{4B5963}
\titleformat{\section}{\large\bfseries\color{ink}}{}{0em}{}
\titleformat{\subsection}{\normalsize\bfseries\color{ink}}{}{0em}{}
\titlespacing*{\section}{0pt}{10pt plus 2pt minus 2pt}{4pt}
\titlespacing*{\subsection}{0pt}{7pt plus 2pt minus 1pt}{3pt}

\makeatletter
\def\@maketitle{\newpage\null\vskip -1.2em%
{\centering{\fontsize{17}{19.5}\selectfont\bfseries\color{ink}\@title\par}\vskip 10pt
}}
\makeatother
\ifLuaTeX
  \usepackage{selnolig}  % disable illegal ligatures
\fi
\IfFileExists{bookmark.sty}{\usepackage{bookmark}}{\usepackage{hyperref}}
\IfFileExists{xurl.sty}{\usepackage{xurl}}{} % add URL line breaks if available
\hypersetup{
  pdftitle={Integrated Design and Reproducible Feasibility Analysis of a Four-Wheel Rover with Packet-Commanded Imaging, LiDAR Ranging, and Learning-Assisted Target Recognition},
  pdfauthor={Sabik Bin Sultan; Shafi Bin Sultan; Safwan Sadad; Aranya Abeer Khan Prapya; Maria Sultan},
  pdflang={en-US},
  colorlinks=true,
  linkcolor={ink},
  filecolor={Maroon},
  citecolor={Blue},
  urlcolor={ink},
  pdfcreator={LaTeX via pandoc}}

\title{Integrated Design and Reproducible Feasibility Analysis of a
Four-Wheel Rover with Packet-Commanded Imaging, LiDAR Ranging, and
Learning-Assisted Target Recognition}
\author{}
\date{}

\begin{document}
\maketitle

\begingroup
\centering\fontsize{9.5}{12}\selectfont
\makebox[\textwidth][c]{%
\textbf{Sabik Bin Sultan}\textsuperscript{1}\quad
\textbf{Shafi Bin Sultan}\textsuperscript{2}\quad
\textbf{Safwan Sadad}\textsuperscript{3}\quad
\textbf{Aranya Abeer Khan Prapya}\textsuperscript{4}\quad
\textbf{Maria Sultan}\textsuperscript{5}}\par
\vspace{3pt}
{\fontsize{9}{11}\selectfont
\textsuperscript{1}BAF Shaheen College Kurmitola\quad
\textsuperscript{2}St. Joseph Higher Secondary School\\
\textsuperscript{3,4}Dhaka Residential Model College\quad
\textsuperscript{5}South Point School \& College\par
\vspace{2pt}
\textsuperscript{1}Team Co-Lead\quad
\textsuperscript{2}Student Team Lead\quad
\textsuperscript{3}Mechanical Lead \& Pilot\\
\textsuperscript{4}Student Safety Officer \& Documentation Lead\quad
\textsuperscript{5}Pilot\par
\vspace{2pt}
\textsuperscript{1}sabikbinsultan@gmail.com\quad
\textsuperscript{2}sj24-11ma-007@sjs.edu.bd\\
\textsuperscript{3}Avoidsafwan@gmail.com\quad
\textsuperscript{4}abeerdrmc@gmail.com\par}
\endgroup
\vspace{5pt}

\textbf{Abstract---} This study develops an integrated, reproducible
design assessment for a four-wheel educational rover intended for the
high-school remote-control division of the 2027 NASA Human Exploration
Rover Challenge. The architecture combines commercial differential drive
control, independently sprung wheel carriers, separate drive and task
batteries, and a Raspberry Pi-based subsystem that executes received
imaging and ranging commands. The analysis reconciles component lists
with dimensioned computer-aided design views, derives mobility and
electrical constraints, and connects sensor geometry to command
execution. At the estimated mass of 10.919 kg, a 215 mm wheel rotating
at 200 RPM produces a theoretical rolling speed of 8.105 km/h;
consequently, the nominal motor label does not establish compliance with
an 8 km/h design ceiling. A steady 20° climb requires at least 1.255 N·m
per wheel under an assumed rolling-resistance coefficient of 0.10 and
equal torque sharing. At 50 ft, an ideal 12-inch feature spans
approximately 71 pixels across a full-resolution Camera Module 3
Standard image, while a 1° pointing error displaces the ranging ray by
approximately 266 mm. An executed synthetic benchmark compares
color--geometry rules and two linear support-vector classifiers on 2,400
generated board crops with scene-disjoint evaluation. A separate
reference command executor is checked using mocked hardware. These
software results and analytical sensitivities support a testable
development plan; the study does not report physical rover trials. The
principal contribution is a traceable connection between component
choices, failure mechanisms, quantitative design limits, and
reproducible verification tasks.


\hypertarget{introduction}{%
\Needspace{4\baselineskip}
\section{1. Introduction}\label{introduction}}

A small rover that can cross uneven terrain and execute a
radio-commanded scientific task presents an integration problem rather
than a collection of independent component-selection problems. Wheel
diameter determines the relationship between motor speed and vehicle
speed. Suspension travel changes load distribution and sensor
orientation. Motor transients can disturb computing and radio reception.
A plausible range value may refer to a background surface rather than
the intended target. These interactions matter particularly when the
same vehicle must meet a competition envelope, operate from batteries,
and present its results immediately after an excursion.

The 2027 NASA Human Exploration Rover Challenge (HERC) remote-control
high-school rules provide the application context
\cite{herc2027}. The rover may be driven into
position, after which the specified task operations are autonomous. In
this paper, \emph{autonomy} means executing received task commands
without manual camera triggering, filter selection, or range
measurement. It does not mean autonomous navigation around the entire
course. This distinction permits a commercial radio-controlled
drivetrain to coexist with a separately programmed task subsystem and
prevents the capability statement from exceeding the proposed
implementation.

The mechanical basis is a student-supplied four-wheel design with a
sloped equipment enclosure, custom printed wheels, and independent
pivoting wheel carriers. The electrical basis comprises four brushed
geared motors, two electronic speed controllers (ESCs), and separate
lithium-polymer batteries for propulsion and task electronics. The task
system combines a Raspberry Pi 5, a camera, a single-beam LiDAR, a
servo-positioned sensor assembly, an onboard display, and a receive-only
software-defined radio chain. The supplied material includes CAD images
and planned component lists, but no native SolidWorks assembly,
calibrated mass properties, motor performance curves, or physical test
logs. The research question is therefore: \textbf{which parts of the
proposed architecture are quantitatively consistent, which limits can be
derived, and which measurements are necessary before the design can be
regarded as demonstrated?}

Three contributions address that question. First, the study reconciles
the latest dimensional and mass estimates with mobility, suspension,
power, and optical calculations. Second, it specifies an executable
command-state architecture and a hardware-independent verification suite
covering command order, requests for fresh captures, fault handling, and
logical state preservation. Third, it evaluates a learning-assisted
target-classification extension on a reproducible synthetic dataset,
with explicit separation between nominal and shifted scene
distributions. The resulting manuscript is a design and software
feasibility study; its experiments concern generated images and mocked
interfaces rather than a completed rover.

\begin{figure}
\centering
\includegraphics[width=6.45in,height=\textheight]{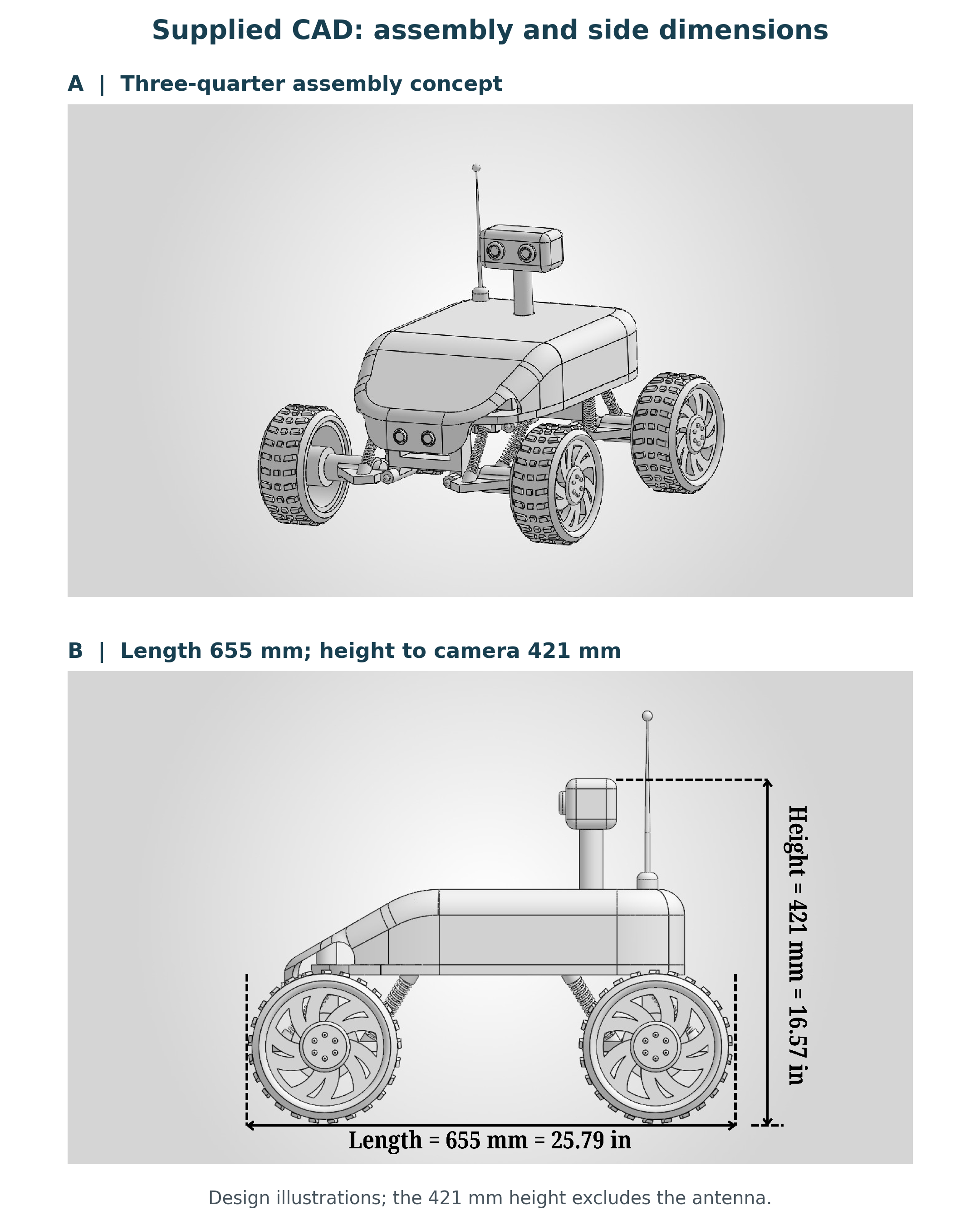}
\caption{Supplied rover CAD views: assembly and side profile. These are
design renderings supplied by the team, not photographs of a fabricated
prototype. The latest dimensioned design takes precedence over earlier
concept images.}
\end{figure}

\hypertarget{evidence-basis-related-work-and-design-requirements}{%
\Needspace{4\baselineskip}
\section{2. Evidence Basis, Related Work, and Design
Requirements}\label{evidence-basis-related-work-and-design-requirements}}

\hypertarget{literature-retrieval-and-scope}{%
\Needspace{3\baselineskip}
\subsection{2.1. Literature retrieval and
scope}\label{literature-retrieval-and-scope}}

A broad discovery search conducted on 15 September 2026 used 20 OpenAlex
queries spanning skid steering, rover suspension, wheel--soil
interaction, printed PETG/TPU structures, camera--range calibration, and
synthetic vision. The queries returned 4,000 metadata occurrences.
Removing 547 duplicates left 3,453 unique records; automated term
matching in titles and available abstracts retained 1,514 topic
candidates and excluded 1,939. This is a documented metadata screen, not
a systematic review or a claim that thousands of full papers were read.
Twenty-five primary research references were selected for closer
relevance assessment: 14 received selective full-text inspection, while
11 were assessed at the available abstract or metadata level. The
accompanying evidence ledger records these access levels and the limits
of each claim.

The search is useful for identifying established failure mechanisms and
suitable analysis methods. It does not justify importing another rover's
friction coefficients, suspension benefits, classification accuracy, or
fatigue life. Search ranking, absent abstracts, and broad keywords also
prevent a defensible claim of exhaustive coverage. Manufacturer
documents and the supplied handbook are treated separately from the
research-paper screen because they establish interface specifications
and task rules rather than independent experimental evidence.

\hypertarget{mobility-suspension-and-printed-wheels}{%
\Needspace{3\baselineskip}
\subsection{2.2. Mobility, suspension, and printed
wheels}\label{mobility-suspension-and-printed-wheels}}

Skid steering requires lateral slip during a turn. Mandow et
al.~identify effective kinematic parameters from measurements, while
Wang et al.~study curvature-dependent instantaneous centers of rotation
on a particular four-wheel platform
\cite{mandow2007,wang2015}. These studies support using
ideal differential-drive equations as an initial command model, followed
by terrain-specific identification. They do not establish universal
turning parameters for the present rover. Dynamic modeling also requires
rolling resistance, terrain shear, actuator saturation, and controller
behavior \cite{yu2010}.

Terramechanics studies provide a basis for investigating wheel sinkage
and traction \cite{iagnemma2004,jia2012,shibly2005}. However, material response,
soil preparation, wheel geometry, and load are necessary inputs. Sutoh
et al.~demonstrate that diameter, mass, and grouser configuration can
alter climbing behavior in their loose-soil tests and that classical
models do not necessarily predict lightweight-rover performance
precisely \cite{sutoh2012}. For this reason,
the present calculations use transparent resistance sweeps rather than
inventing a calibrated soil model.

The supplied suspension is also distinct from the adaptive rocker
concept discussed by Reina and Foglia
\cite{reina2010}. It uses independent pivot
arms and paired springs, rather than a rocker-type mechanism with
variable wheel-camber compensation. Work on spring-suspended skid-steer
rovers motivates considering suspension-dependent wheel contact geometry
\cite{martinez2010}, while the Mars
Exploration Rover mobility literature illustrates the separate
importance of design and physical verification
\cite{lindemann2005}. Neither source makes
the current CAD assembly a flight-qualified or terrain-validated
mechanism.

Printed-wheel literature supports a component-level characterization
program. Wang et al.~manufactured TPU specimens and non-pneumatic tires
and reported sensitivity to manufacturing conditions
\cite{wang2020}. PETG studies similarly show
that printed response depends on processing parameters and reinforcement
\cite{hsueh2021,szykiedans2017}. Competition-rover
wheels with separately printed rims and compliant tires provide a
fabrication precedent \cite{zakrecki2018}.
More recent PETG/TPU multi-material work and PLA/TPU topology studies
broaden the design space \cite{zagorski2026,dezianian2023}, but their material
systems and measured properties cannot be assigned directly to the
proposed unfilled PETG rim and TPU 95A tire.

\hypertarget{sensing-and-learning}{%
\Needspace{3\baselineskip}
\subsection{2.3. Sensing and learning}\label{sensing-and-learning}}

Camera calibration establishes the mapping between pixels and viewing
rays. Zhang's planar-target method provides an established foundation
for identifying camera intrinsics and distortion
\cite{zhang2000calibration}. Camera--range
registration is a separate problem: approaches using multiple
checkerboards or alignment of camera and laser features demonstrate the
importance of overlapping geometry and independent calibration
\cite{geiger2012single,levinson2013online}. Those methods were
developed for richer range observations than the single beam available
here. The present design therefore requires an instrumented, multi-pose
calibration procedure adapted to a single-beam sensor.

Synthetic-data studies motivate controlled variation of lighting,
texture, pose, and background
\cite{tobin2017domain,tremblay2018synthetic,richter2016playing}. They do not imply
that a classifier trained on simple generated boards will transfer to
outdoor camera images. Dynamics randomization concerns physical/control
variation \cite{peng2018dynamics}, while
domain adaptation can combine simulated and real visual experience
\cite{bousmalis2018grasping}. These are
distinct mechanisms. Terrain classification using proprioceptive
measurements is another relevant rover application of machine learning
\cite{dimastrogiovanni2020}, but the present
experiment is restricted to visual classification of a supplied board
crop.

\hypertarget{traceable-design-basis}{%
\Needspace{3\baselineskip}
\subsection{2.4. Traceable design basis}\label{traceable-design-basis}}

The latest technical specification is used where earlier drafts
conflict. In particular, the camera-height dimension is 421 mm, the rim
diameter is 188 mm, and the tire radial thickness is 13.5 mm. Earlier
values of 432 mm, 185 mm, and 15 mm are not mixed into the calculations.
The 421 mm height excludes the antenna; a complete excursion-ready
envelope still requires a measurement that includes every installed
projection.

\FloatBarrier
\begingroup\small
\begin{longtable}[]{@{}
  >{\raggedright\arraybackslash}p{(\columnwidth - 4\tabcolsep) * \real{0.3333}}
  >{\raggedright\arraybackslash}p{(\columnwidth - 4\tabcolsep) * \real{0.3333}}
  >{\raggedright\arraybackslash}p{(\columnwidth - 4\tabcolsep) * \real{0.3333}}@{}}
\caption{Adopted design quantities and their evidence
status.}\tabularnewline
\toprule\noalign{}
\begin{minipage}[b]{\linewidth}\raggedright
Quantity
\end{minipage} & \begin{minipage}[b]{\linewidth}\raggedright
Adopted value or interpretation
\end{minipage} & \begin{minipage}[b]{\linewidth}\raggedright
Evidence status
\end{minipage} \\
\midrule\noalign{}
\endfirsthead
\toprule\noalign{}
\begin{minipage}[b]{\linewidth}\raggedright
Quantity
\end{minipage} & \begin{minipage}[b]{\linewidth}\raggedright
Adopted value or interpretation
\end{minipage} & \begin{minipage}[b]{\linewidth}\raggedright
Evidence status
\end{minipage} \\
\midrule\noalign{}
\endhead
\bottomrule\noalign{}
\endlastfoot
Overall length × width & 655 × 630 mm & Supplied design dimensions \\
Height to camera top & 421 mm, antenna excluded & Supplied design
dimension \\
Wheels & Four; 215 mm diameter × 90 mm width & Supplied design
dimensions \\
Rim / tire radial thickness & 188 mm / 13.5 mm & Dimension closure
verified \\
Estimated fully equipped mass & 10.919 kg & Team-supplied theoretical
estimate; unweighed \\
Track / wheelbase & 540 / 440 mm & Conditional inference from outer
dimensions \\
Motors & Four 12 V brushed geared units, nominally about 200 RPM &
Listed class; continuous curves unavailable \\
Drive / task battery & 3S, 5,000 / 2,200 mAh & Planned components \\
Task computer / camera & Raspberry Pi 5, 8 GB / Camera Module 3 Standard
& Planned components \\
Range sensor & TF03-100 with serial interface & Planned component \\
Camera pan mechanism & Servo-positioned, bearing-supported platform &
Proposed; usable angular range unverified \\
\end{longtable}
\endgroup

The inferred track assumes that the stated width is the outer tire width
and that each tire is symmetrically located: 630 − 90 = 540 mm. The
inferred wheelbase assumes that the front and rear wheel extrema
establish overall length: 655 − 215 = 440 mm. These are conditional
contact-geometry estimates, not extracted native-CAD measurements.

\FloatBarrier
\begingroup\small
\begin{longtable}[]{@{}
  >{\raggedright\arraybackslash}p{(\columnwidth - 4\tabcolsep) * \real{0.3333}}
  >{\raggedright\arraybackslash}p{(\columnwidth - 4\tabcolsep) * \real{0.3333}}
  >{\raggedright\arraybackslash}p{(\columnwidth - 4\tabcolsep) * \real{0.3333}}@{}}
\caption{Handbook requirements mapped to verification
needs.}\tabularnewline
\toprule\noalign{}
\begin{minipage}[b]{\linewidth}\raggedright
Handbook consideration
\end{minipage} & \begin{minipage}[b]{\linewidth}\raggedright
Design interpretation
\end{minipage} & \begin{minipage}[b]{\linewidth}\raggedright
Evidence needed before acceptance
\end{minipage} \\
\midrule\noalign{}
\endfirsthead
\toprule\noalign{}
\begin{minipage}[b]{\linewidth}\raggedright
Handbook consideration
\end{minipage} & \begin{minipage}[b]{\linewidth}\raggedright
Design interpretation
\end{minipage} & \begin{minipage}[b]{\linewidth}\raggedright
Evidence needed before acceptance
\end{minipage} \\
\midrule\noalign{}
\endhead
\bottomrule\noalign{}
\endlastfoot
Excursion-ready cube, Rule 5.5.4 & 762 mm on each axis & Measure
complete rover, including antenna \\
High-school manufacture, Rule 6.4.2.1 & Team-designed/manufactured
chassis, wheels and tires; permitted commercial drive/suspension parts &
Manufacturing records and final configuration \\
Primary drive control, Rule 6.4.3 & Commercial transmitter, receiver and
ESC control & Demonstration of configured control and failsafe \\
Speed, Rule 6.4.4 & Use 8 km/h as the conservative ceiling associated
with the stated 5 mph limit & Full-charge maximum-speed test and limiter
verification \\
Mass, Rules 6.4.5 and Section 8.2 & 60 lb absolute limit; 20 lb is a
scoring objective & Weigh excursion-ready vehicle \\
Battery protection, Rules 6.4.6--6.4.7 & Separate insulated enclosures;
main overcurrent protection & Wiring and enclosure inspection \\
Task autonomy, Rule 6.4.8 & Manual positioning followed by automatic
task execution & Complete received-command sequence demonstration \\
\end{longtable}
\endgroup

These interpretations refer to the supplied 2027 handbook
\cite{herc2027}. The calculated mass is 24.072
lb, below the stated 60 lb absolute limit but above the 20 lb scoring
objective. Reaching 20 lb would require removing 1.847 kg, or 16.92\% of
the present estimate. The calculation is a mass-management target, not a
reason to remove load-bearing material without re-evaluating strength.

\hypertarget{integrated-mechanical-and-drive-architecture}{%
\Needspace{4\baselineskip}
\section{3. Integrated Mechanical and Drive
Architecture}\label{integrated-mechanical-and-drive-architecture}}

\hypertarget{chassis-wheel-carriers-and-load-paths}{%
\Needspace{3\baselineskip}
\subsection{3.1. Chassis, wheel carriers, and load
paths}\label{chassis-wheel-carriers-and-load-paths}}

The proposed chassis uses 6061-T6 aluminum tube and sheet to support the
equipment enclosure, batteries, suspension pivots, and sensor mast.
Bolted connections provide assembly access and allow individual modules
to be replaced. The battery masses should be placed low and secured
against movement, because a loose pack changes both electrical
reliability and center of mass. The enclosure must permit cooling
airflow and service access without interfering with wheel travel or
exposed moving joints.

Each wheel carrier pivots relative to the chassis and carries its motor,
spindle supports, wheel, and spring attachments. Keeping the motor and
spindle on the same moving carrier preserves their relative alignment as
the suspension articulates. A coupling connects the nominal 6 mm motor
shaft to the wheel spindle. The wheel spindle is supported by two
separated bearings and positively retained at the hub. This load path is
intended to transfer wheel contact forces through the spindle bearings
and carrier rather than assuming that a small gearbox output bearing can
sustain terrain impacts.

The CAD concept shows two coils at each wheel, with one damper
associated with one of the coils. Thus the complete rover has eight
springs and four dampers. The architecture should not be described as
having two dampers per wheel. Before fabrication, the final assembly
must establish pivot-axis alignment, bearing fits, end stops, spring
seating, cable bending radii, and full-travel interference clearances.
Screenshots cannot reveal interference through every degree of freedom
or substitute for an assembly-motion study.

\begin{figure}
\centering
\includegraphics[width=6.45in,height=\textheight]{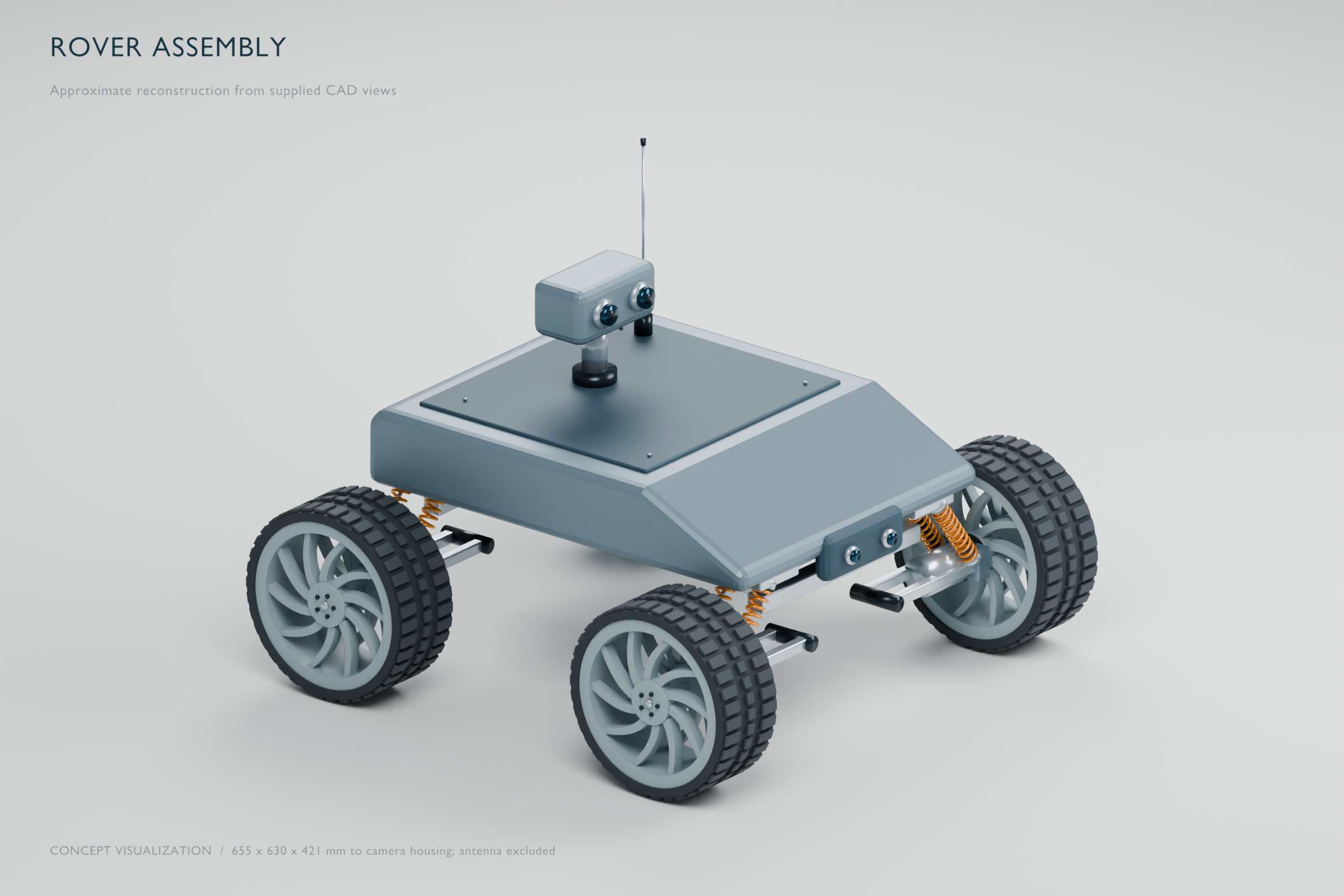}
\caption{New Blender reconstruction of the proposed rover. Geometry is
an approximate reconstruction from the supplied CAD views and
dimensions; the rendering illustrates packaging and appearance, not a
manufactured vehicle or structural simulation.}
\end{figure}

\hypertarget{wheel-manufacture-and-structural-verification}{%
\Needspace{3\baselineskip}
\subsection{3.2. Wheel manufacture and structural
verification}\label{wheel-manufacture-and-structural-verification}}

The outer wheel dimension closes geometrically as

\[D_o=D_r+2t=188+2(13.5)=215\ \mathrm{mm}.\]

The proposed PETG rim provides a relatively stiff hub and spoke
structure, while the TPU 95A tire supplies a compliant tread. ``95A''
identifies a nominal hardness class; it does not establish the
stiffness, fatigue resistance, or tear strength of the printed tire. The
manufacturing record should therefore include filament identity, drying
procedure, extrusion temperature, layer height, wall count, infill
topology and fraction, print orientation, and any bonding or mechanical
interlock between rim and tire. The printed interface requires its own
retention check.

A defensible strength analysis begins with final geometry and measured
or appropriately conservative material properties. For a spindle
idealized as a solid, circular, linearly elastic shaft under bending and
torsion, bending and torsion give

\[\sigma_b=\frac{32M}{\pi d^3},\qquad \tau_t=\frac{16T}{\pi d^3},\qquad
\sigma_{\mathrm{vm}}=\sqrt{\sigma_b^2+3\tau_t^2}.\]

These expressions require the bearing span, overhang, bending moment,
applied torque, and actual spindle diameter. None of those missing
inputs is inferred from the visual thickness of a rendered part. Bearing
reactions must be computed from the real support locations; a hub
overhang can increase bearing reactions beyond the simple wheel-load
value. Thread roots, keyways, printed notches, and joint preload need
local treatment rather than a smooth-shaft formula alone.

Printed coupons should first characterize the selected process in the
relevant directions. Rim/hub tests should then apply radial, lateral,
and torsional loads representative of the assembly. Tire tests should
measure load--deflection behavior, permanent set, traction on the
intended surfaces, and tread or interface damage after repeated use. A
finite-element model can subsequently be correlated to these
measurements. No stress contour or safety factor is presented here
because the required native geometry, boundary conditions, and material
characterization were not supplied.

\begin{figure}
\centering
\includegraphics[width=3.45in,height=\textheight]{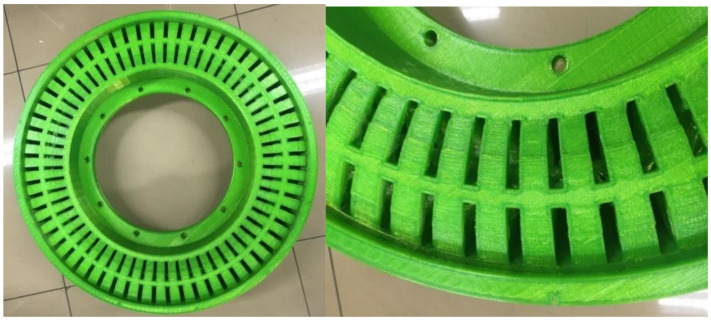}
\caption{Published FDM TPU tire used as a manufacturing comparison.
Reproduced from J. Wang et al., original Fig. 12
\cite{wang2020}, under CC BY 4.0
(https://creativecommons.org/licenses/by/4.0/). This is the cited
authors' tire, not a photograph of the present rover wheel; its
properties are not transferred to this design.}
\end{figure}

\hypertarget{commercial-differential-drive-and-electrical-connections}{%
\Needspace{3\baselineskip}
\subsection{3.3. Commercial differential drive and electrical
connections}\label{commercial-differential-drive-and-electrical-connections}}

The transmitter mixes throttle and steering into two centered command
channels. The receiver sends one pulse-width control signal to the left
ESC and one to the right ESC, with their required signal returns. Each
reversible brushed ESC drives the two motors on its side in parallel.
Forward and reverse operation therefore change the polarity across the
motor pair; neither motor output should be treated as a permanent
chassis ground. Motor direction is checked with the wheels raised before
the rover is allowed to move.

Only one ESC battery-eliminator circuit (BEC) supplies the receiver. The
positive BEC conductor from the other ESC is isolated while its signal
and reference return remain correctly connected. This avoids assuming
that two independent regulated outputs can be paralleled. Receiver
failsafe and transmitter-loss behavior must command neutral on both
sides, and the test must include loss of radio reception under motor
load. Neutral pulse values and ESC calibration are specific to the
selected hardware.

The drive pack feeds an accessible disconnect and a proposed 20 A main
fuse. A proposed 5 A time-delay fuse is included in one lead of each
individual motor branch. These numbers are provisional coordination
choices. Their sum is not a calculation of operating current, and a
label rating does not establish interruption capacity, transient
tolerance, wire protection, or selectivity. Motor-start, turn, stall,
and sustained-climb currents must be measured with the final wiring and
load.

\begin{figure}
\centering
\includegraphics[width=6.45in,height=\textheight]{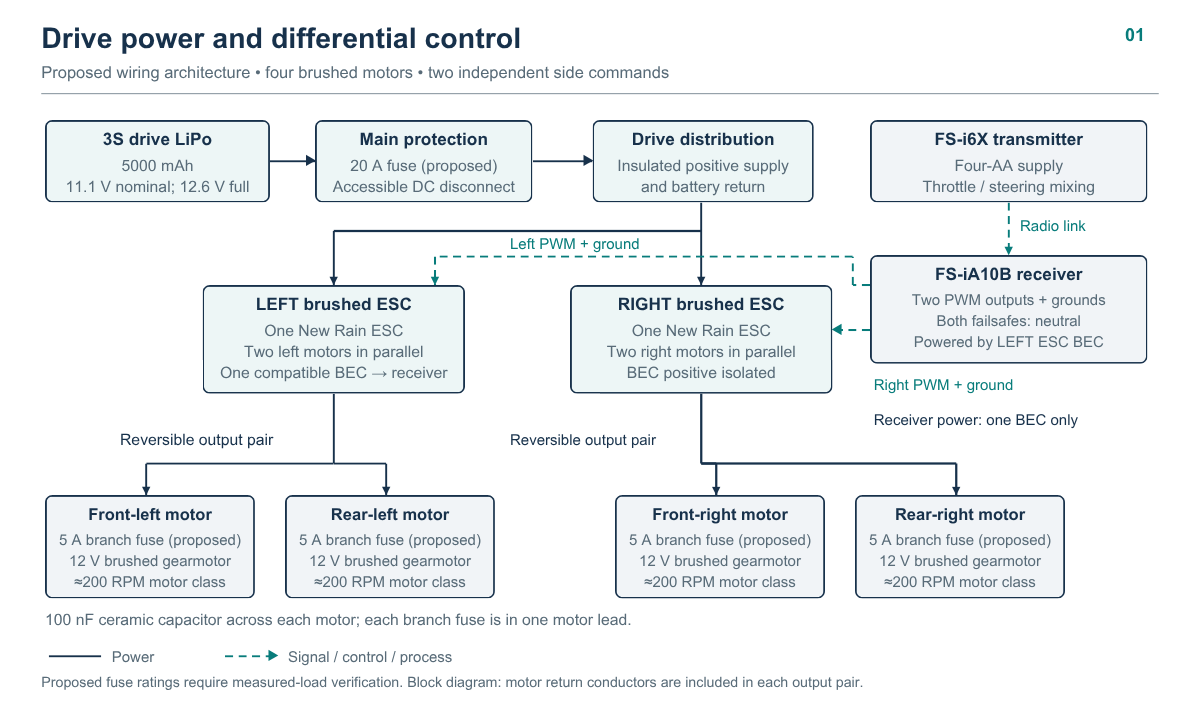}
\caption{Editable Canva drive-system block diagram. It separates
receiver commands, the single active BEC supply, battery distribution,
and reversible motor outputs. Each motor-output connection represents a
conductor pair; detailed connector pinouts remain a wiring-drawing
task.}
\end{figure}

For normalized throttle \(u\) and steering \(s\), an initial mixer is

\[a=u-s,\quad b=u+s,\quad q=\max(1,|a|,|b|),\quad
u_L=a/q,\quad u_R=b/q.\]

This jointly normalized mixer is a conceptual allocation rule. Its
implementation must be checked against the commercial transmitter's
actual mixing and saturation behavior; independent channel clipping does
not preserve the same command ratio. Joint normalization retains the
left/right ratio while respecting the allowed command range. Sign
conventions must be verified against the installed motor orientation.
This is a mixing relationship, not a calibrated mapping from stick
position to ground speed.

Ideal differential-drive kinematics give

\[v_x=\frac{r}{2}(\omega_R+\omega_L),\qquad
\dot{\psi}=\frac{r}{T_{\mathrm{eff}}}(\omega_R-\omega_L).\]

Here \(\omega_L\) and \(\omega_R\) are wheel angular speeds in rad/s.
The effective separation \(T_{\mathrm{eff}}\) equals geometric track
only under the idealized model. In actual skid steering it depends on
slip, terrain, curvature, wheel loading, and suspension configuration
\cite{mandow2007,wang2015}. Changing effective separation
models yaw sensitivity but does not itself compensate longitudinal slip.
A pivot turn is therefore a demanding current and traction case, not a
``zero-radius'' capability established by the wiring diagram. The
proposed drivetrain does not include a demonstrated encoder-based
autonomous velocity controller.

\begin{figure}
\centering
\includegraphics[width=3.85in,height=\textheight]{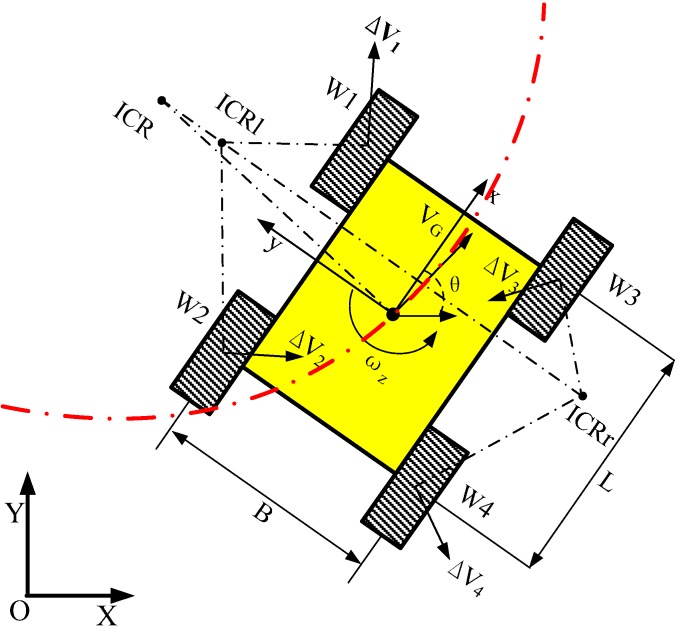}
\caption{Published skid-steer instantaneous-center geometry, reproduced
from T. Wang et al., original Fig. 1
\cite{wang2015}, under CC BY 4.0
(https://creativecommons.org/licenses/by/4.0/). The symbols and geometry
belong to that paper. The figure motivates measured slip-aware
kinematics; no fitted coefficient is copied into the present rover
model.}
\end{figure}

\hypertarget{mobility-traction-and-stability-analysis}{%
\Needspace{4\baselineskip}
\section{4. Mobility, Traction, and Stability
Analysis}\label{mobility-traction-and-stability-analysis}}

\hypertarget{wheel-speed-and-the-full-charge-case}{%
\Needspace{3\baselineskip}
\subsection{4.1. Wheel speed and the full-charge
case}\label{wheel-speed-and-the-full-charge-case}}

For wheel diameter \(D\) in meters and rotational speed \(n\) in RPM,
ideal rolling speed is

\[v_{\mathrm{km/h}}=\frac{\pi Dn\,60}{1000}.\]

At \(D=0.215\) m, 190 RPM gives 7.70004 km/h and 200 RPM gives 8.10531
km/h. The 8 km/h boundary is

\[n_{8}=\frac{8(1000)}{\pi(0.215)(60)}=197.40148\ \mathrm{RPM}.\]

Consequently, a nominal ``200 RPM'' motor description does not
demonstrate compliance. The earlier assumption that loading will reduce
speed to 190 RPM is a scenario, not a measurement. A light-load,
high-voltage condition may be the relevant maximum-speed case.

If, and only if, a motor actually runs at 200 RPM at 12 V with
approximately linear no-load voltage scaling, a fully charged 3S pack at
12.6 V would imply 210 RPM and 8.51057 km/h. The ratio 8/8.51057 is
approximately 0.940. This is a first-pass endpoint estimate, not a
verified ESC limiter setting: PWM behavior, no-load specifications, tire
deformation, battery sag, and terrain all affect the result. A lower
provisional target, followed by a measured full-charge test over a known
distance, provides room for calibration uncertainty. The limiter must be
assessed in every permitted operating mode, including reverse if
relevant.

\begin{figure}
\centering
\includegraphics[width=6.25in,height=\textheight]{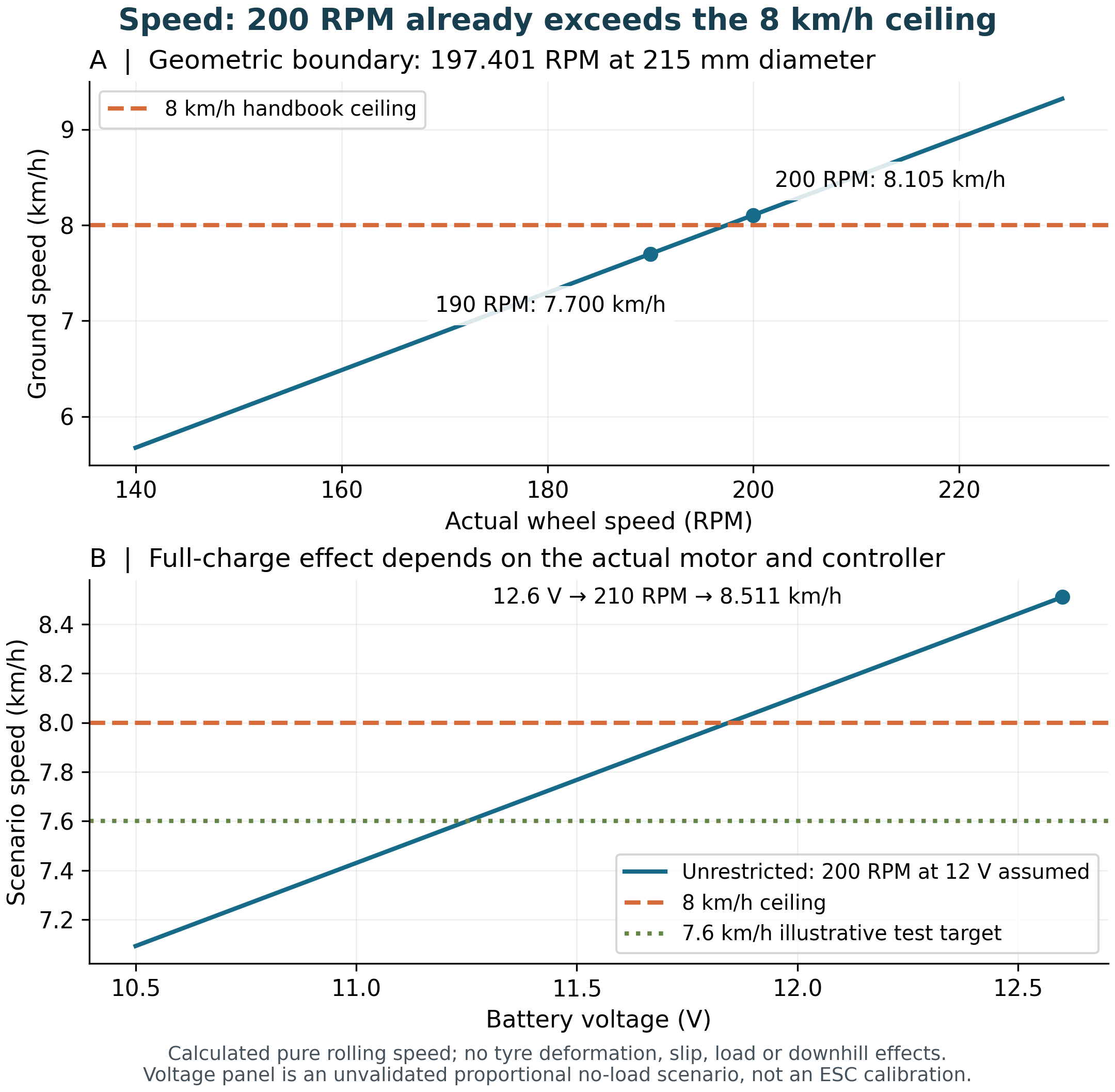}
\caption{Calculated rolling speed versus wheel RPM and an explicitly
conditional voltage-scaling scenario. These curves use the nominal 215
mm diameter; they are not measured speed traces.}
\end{figure}

\hypertarget{straight-climb-lower-bounds}{%
\Needspace{3\baselineskip}
\subsection{4.2. Straight-climb lower
bounds}\label{straight-climb-lower-bounds}}

Let \(m\) denote total mass, \(g\) gravitational acceleration,
\(\theta\) slope angle, and \(C_{rr}\) an assumed rolling-resistance
coefficient. A steady straight climb with no acceleration or additional
soil deformation is approximated by

\[F_{\mathrm{req}}=mg(\sin\theta+C_{rr}\cos\theta),\qquad
\tau_w=\frac{rF_{\mathrm{req}}}{4}.\]

The equal-sharing expression gives the required output torque at each
wheel. Motor-shaft torque before a gearbox is a different quantity. With
available tire--terrain friction coefficient \(\mu\), the corresponding
traction condition is

\[\mu mg\cos\theta\geq F_{\mathrm{req}},\qquad
\mu\geq\tan\theta+C_{rr}.\]

This is a necessary global traction condition, not a sufficient
per-wheel condition. Grade-dependent load transfer and unequal contact
loads can cause an individual wheel to slip even when the aggregate
inequality is satisfied.

For \(m=10.919\) kg, \(r=0.1075\) m, \(g=9.80665\) m/s\textsuperscript{2}, and
\(C_{rr}=0.10\), the resulting values are as follows.

\FloatBarrier
\begingroup\small
\begin{longtable}[]{@{}
  >{\raggedleft\arraybackslash}p{(\columnwidth - 6\tabcolsep) * \real{0.2500}}
  >{\raggedleft\arraybackslash}p{(\columnwidth - 6\tabcolsep) * \real{0.2500}}
  >{\raggedleft\arraybackslash}p{(\columnwidth - 6\tabcolsep) * \real{0.2500}}
  >{\raggedleft\arraybackslash}p{(\columnwidth - 6\tabcolsep) * \real{0.2500}}@{}}
\caption{Calculated steady straight-climb lower bounds at assumed
rolling resistance 0.10.}\tabularnewline
\toprule\noalign{}
\begin{minipage}[b]{\linewidth}\raggedleft
Slope
\end{minipage} & \begin{minipage}[b]{\linewidth}\raggedleft
Required total force (N)
\end{minipage} & \begin{minipage}[b]{\linewidth}\raggedleft
Wheel-output torque per wheel (N·m)
\end{minipage} & \begin{minipage}[b]{\linewidth}\raggedleft
Minimum friction coefficient
\end{minipage} \\
\midrule\noalign{}
\endfirsthead
\toprule\noalign{}
\begin{minipage}[b]{\linewidth}\raggedleft
Slope
\end{minipage} & \begin{minipage}[b]{\linewidth}\raggedleft
Required total force (N)
\end{minipage} & \begin{minipage}[b]{\linewidth}\raggedleft
Wheel-output torque per wheel (N·m)
\end{minipage} & \begin{minipage}[b]{\linewidth}\raggedleft
Minimum friction coefficient
\end{minipage} \\
\midrule\noalign{}
\endhead
\bottomrule\noalign{}
\endlastfoot
0° & 10.708 & 0.288 & 0.100 \\
10° & 29.139 & 0.783 & 0.276 \\
15° & 38.057 & 1.023 & 0.368 \\
20° & 46.685 & 1.255 & 0.464 \\
25° & 54.958 & 1.477 & 0.566 \\
30° & 62.813 & 1.688 & 0.677 \\
\end{longtable}
\endgroup

At 20° and 2 km/h, wheel mechanical power is \(Fv=25.936\) W. If the
total battery-to-wheel efficiency were 0.60, propulsion input would be
43.227 W. This efficiency is an illustrative assumption, not a measured
motor/ESC map. Acceleration adds \(ma\) to longitudinal force; turning,
sinkage, wheel unloading, impacts, and drivetrain losses can increase
demand further. The tabulated torques are therefore lower bounds before
design margin, not sufficient procurement specifications by themselves.

The continuous torque--speed--current curve is needed to determine
whether each motor can sustain the required output at the desired speed.
Stall torque is not an acceptable substitute for continuous torque.
Similarly, a high advertised ESC peak rating does not establish reliable
low-speed operation under repeated skid turns. The mass-reduced
comparison in the figure shows the direct effect of mass on the
simplified force model, while preserving the uncertainty in terrain
parameters.

\begin{figure}
\centering
\includegraphics[width=5.8in,height=\textheight]{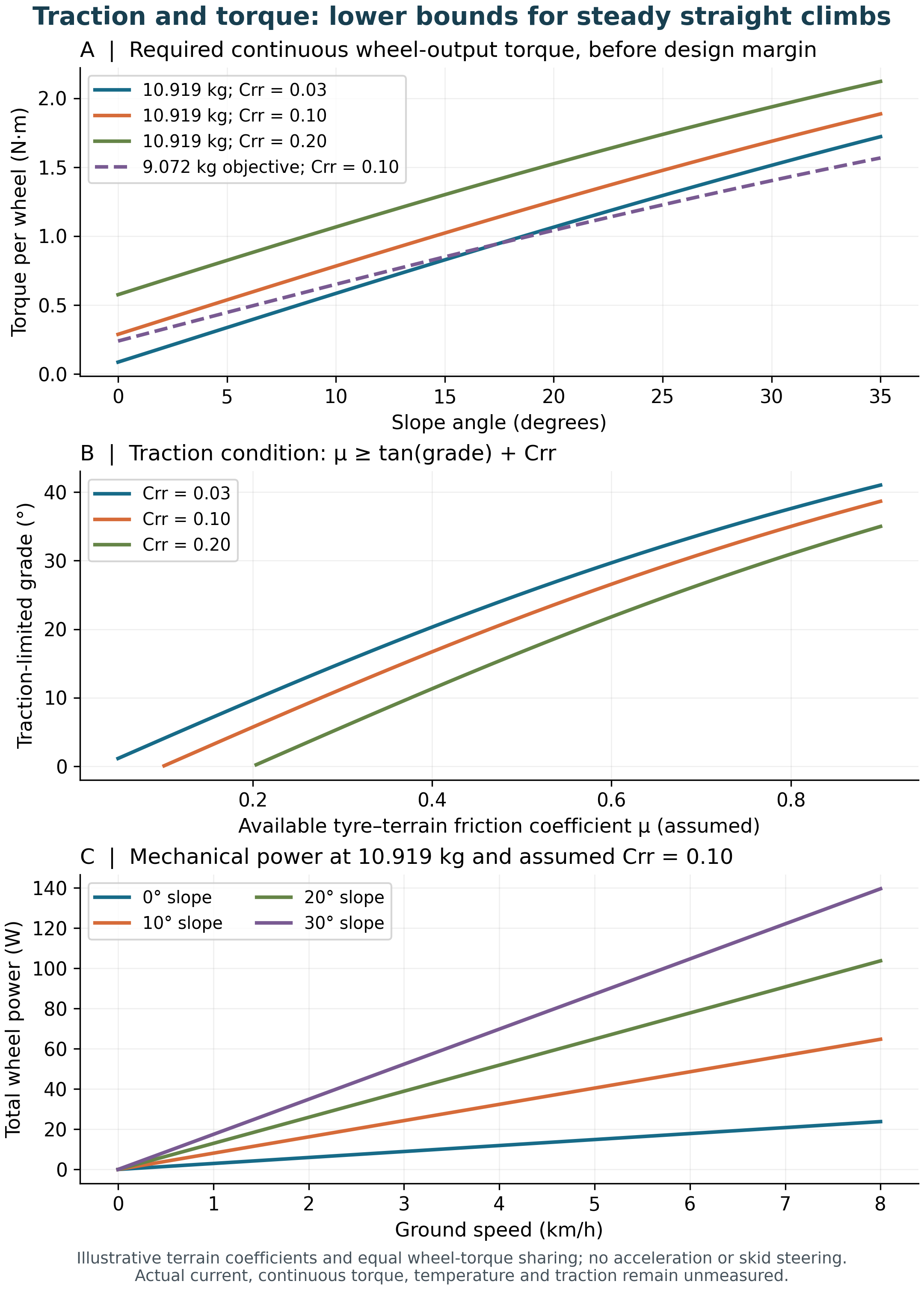}
\caption{Analytical straight-climb sensitivities: wheel-output torque,
friction-limited grade, and mechanical power. Rolling resistance, equal
torque sharing and steady motion are declared assumptions. Turning and
transient losses are excluded.}
\end{figure}

\hypertarget{static-stability-and-changing-contact-geometry}{%
\Needspace{3\baselineskip}
\subsection{4.3. Static stability and changing contact
geometry}\label{static-stability-and-changing-contact-geometry}}

For a centered center of mass at height \(h\) above an assumed
rectangular support plane, first-order lateral and longitudinal tip
angles are

\[\theta_{\mathrm{lat}}=\tan^{-1}\left(\frac{T}{2h}\right),\qquad
\theta_{\mathrm{long}}=\tan^{-1}\left(\frac{L}{2h}\right).\]

Using the conditional \(T=0.540\) m and \(L=0.440\) m with an
illustrative \(h=0.200\) m yields 53.47° and 47.73°, respectively. These
are geometric thresholds for that idealized configuration, not approved
operating slopes. The actual center of mass is unknown; suspension
deflection, terrain contact changes, lateral acceleration, tire
deformation, and payload movement can reduce available margin. If one
wheel loses contact, the support polygon may become triangular and must
be recomputed.

A useful first measurement is the loaded corner weight on a level
surface. Four equal shares would each be 26.770 N, but the actual shares
should be recorded with the installed batteries and task hardware.
Controlled tilt measurements or a validated mass-property model can then
locate the center of mass. The curve family below is more honest than
reporting one ``maximum slope'' from an assumed height.

\begin{figure}
\centering
\includegraphics[width=6.25in,height=\textheight]{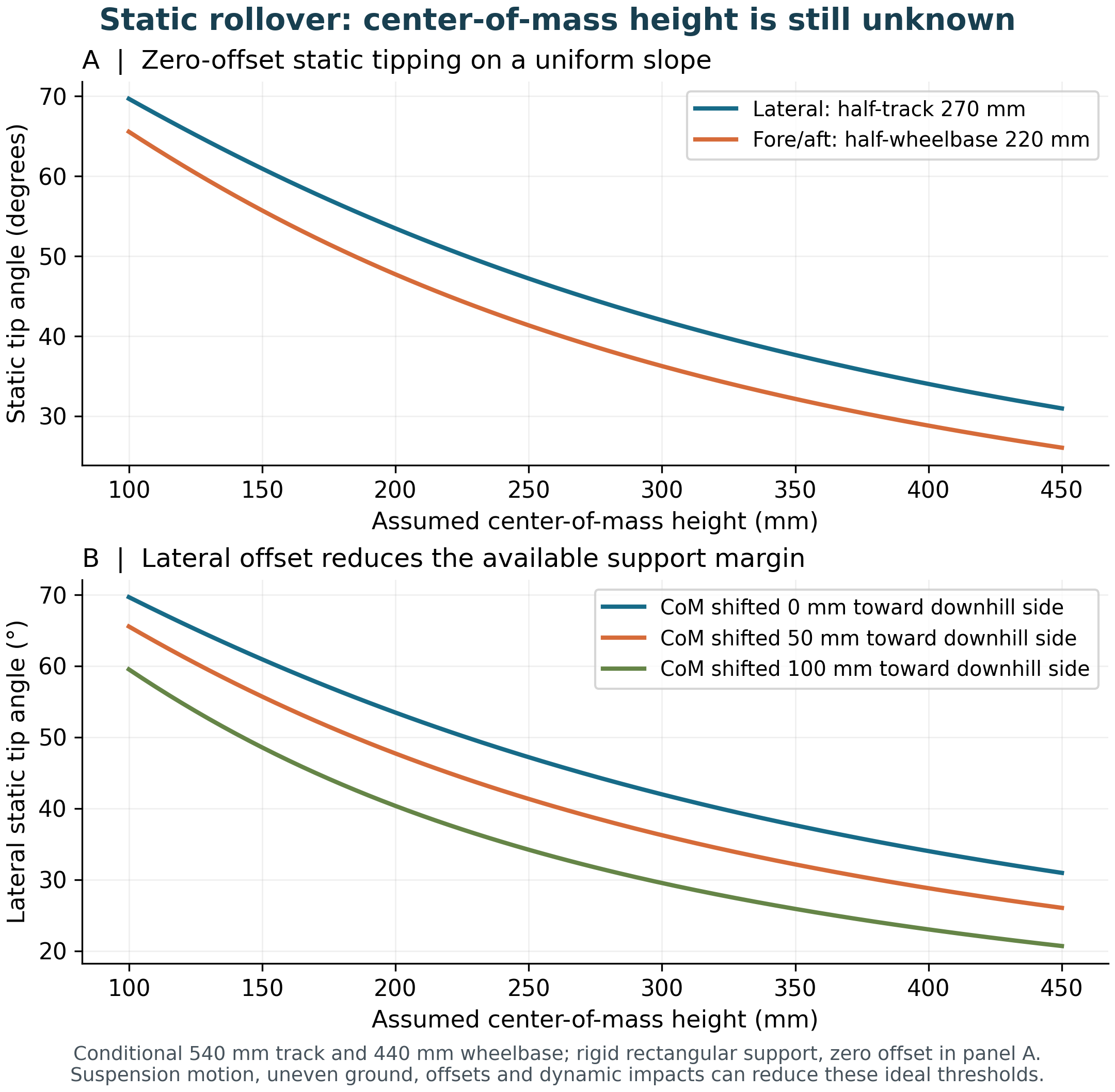}
\caption{Conditional static tip-angle sensitivity to center-of-mass
height. The support rectangle is inferred from the supplied external
dimensions; the center-of-mass height is swept rather than measured.}
\end{figure}

\hypertarget{independent-suspension-and-structural-integration}{%
\Needspace{4\baselineskip}
\section{5. Independent Suspension and Structural
Integration}\label{independent-suspension-and-structural-integration}}

\hypertarget{local-wheel-rate-and-damping}{%
\Needspace{3\baselineskip}
\subsection{5.1. Local wheel rate and
damping}\label{local-wheel-rate-and-damping}}

For small wheel displacement \(z\) near a chosen ride height, let
\(\rho_i=dx_i/dz\) be the local motion ratio between spring compression
and wheel displacement. With two springs of axial stiffness \(k_1\) and
\(k_2\), the local wheel rate is approximately

\[k_w=k_1\rho_1^2+k_2\rho_2^2.\]

This expression assumes locally constant motion ratios. If linkage
geometry changes significantly or spring preload creates a geometric
stiffness contribution, a full derivative of force with respect to wheel
displacement is required. The installation angle alone does not
generally define the motion ratio; attachment lever arms also matter.
Two springs in the drawing cannot simply be treated as twice the wheel
stiffness of one upright spring.

With one damper of axial coefficient \(c_d\) and motion ratio
\(\rho_d\), the local wheel damping is \(c_w=c_d\rho_d^2\). For a
linearized sprung corner mass \(m_s\),

\[f_n=\frac{1}{2\pi}\sqrt{\frac{k_w}{m_s}},\qquad
\zeta=\frac{c_w}{2\sqrt{k_wm_s}}.\]

The analysis sweeps spring rates and sprung masses because these
quantities are not established by the purchase list. A total
quarter-vehicle mass is not automatically a sprung mass: wheels,
carriers, and portions of the motor assembly move differently from the
chassis. The figures are sensitivity tools for selecting and measuring
components, not predicted ride traces.

\begin{figure}
\centering
\includegraphics[width=6.45in,height=\textheight]{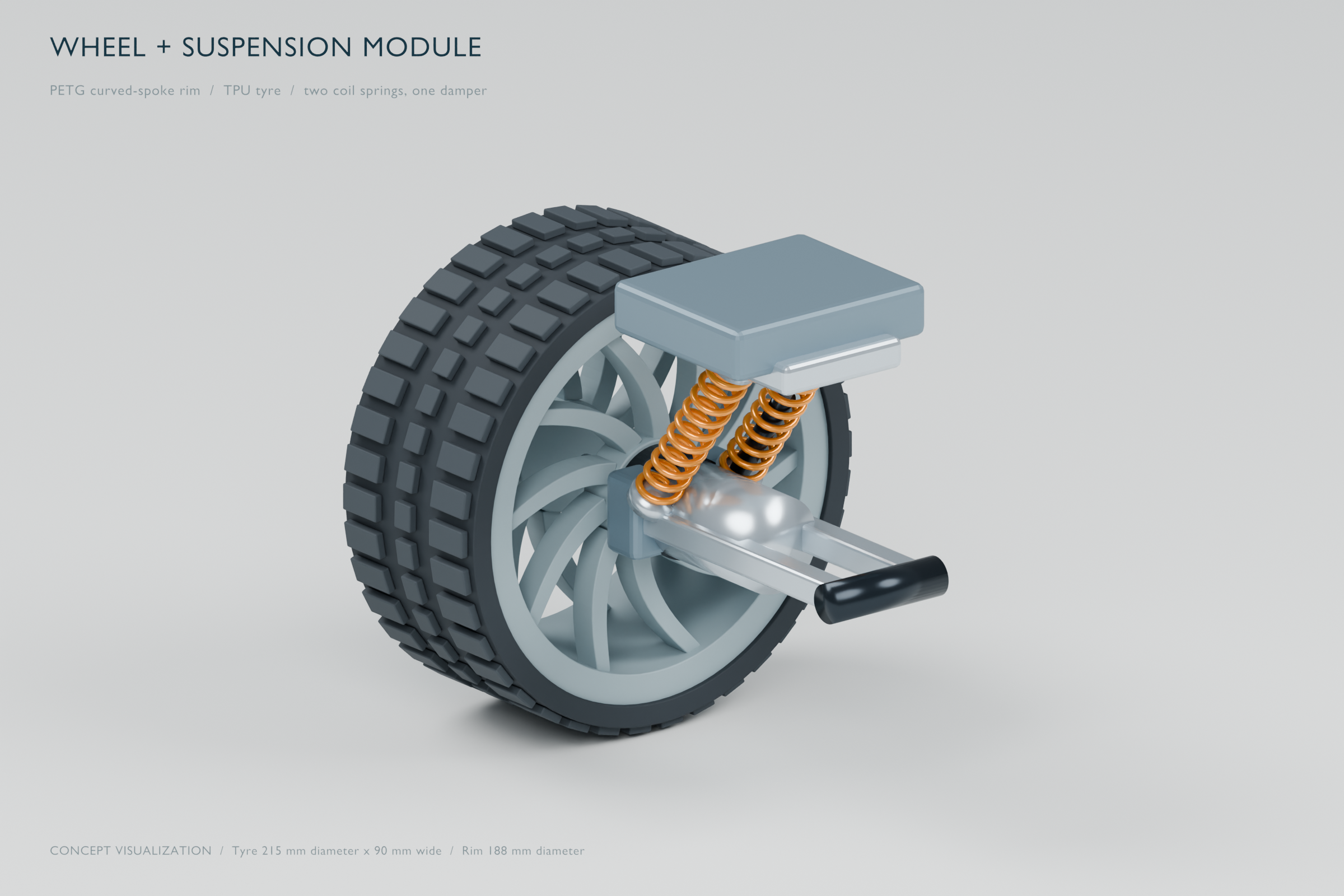}
\caption{Blender illustration of one wheel module, showing the paired
coils and single damper arrangement. The scene reconstructs the visible
mechanism; spring rate, damping coefficient, bearing fits and full
suspension travel remain to be measured or finalized.}
\end{figure}

\hypertarget{integration-checks-and-practical-consequences}{%
\Needspace{3\baselineskip}
\subsection{5.2. Integration checks and practical
consequences}\label{integration-checks-and-practical-consequences}}

The suspension should be tested first as a single loaded corner. Measure
wheel displacement, spring compression, force, and return behavior over
the intended travel. This reveals motion ratios, preload, friction,
binding, and end-stop behavior without requiring a complete vehicle. A
static test can estimate the wheel-rate curve, but damping needs a
dynamic measurement. Recording free decay after a small displacement is
useful only when the response is sufficiently linear and the sensors can
resolve the motion.

At full assembly level, compare wheel loads and ground clearance at
nominal ride height, one-wheel elevation, and diagonal articulation. The
sensor mast moves with the chassis, so roll and pitch alter the camera
and LiDAR viewing geometry. The task system should therefore perform its
precise measurements from a settled position. Passive springs do not
provide active leveling, and the present drawings do not establish an
obstacle-height capability. Tire clearance must be checked at both
compression and rebound, including wiring, fuse holders, and motor
connectors on the moving carriers.

\begin{figure}
\centering
\includegraphics[width=5.8in,height=\textheight]{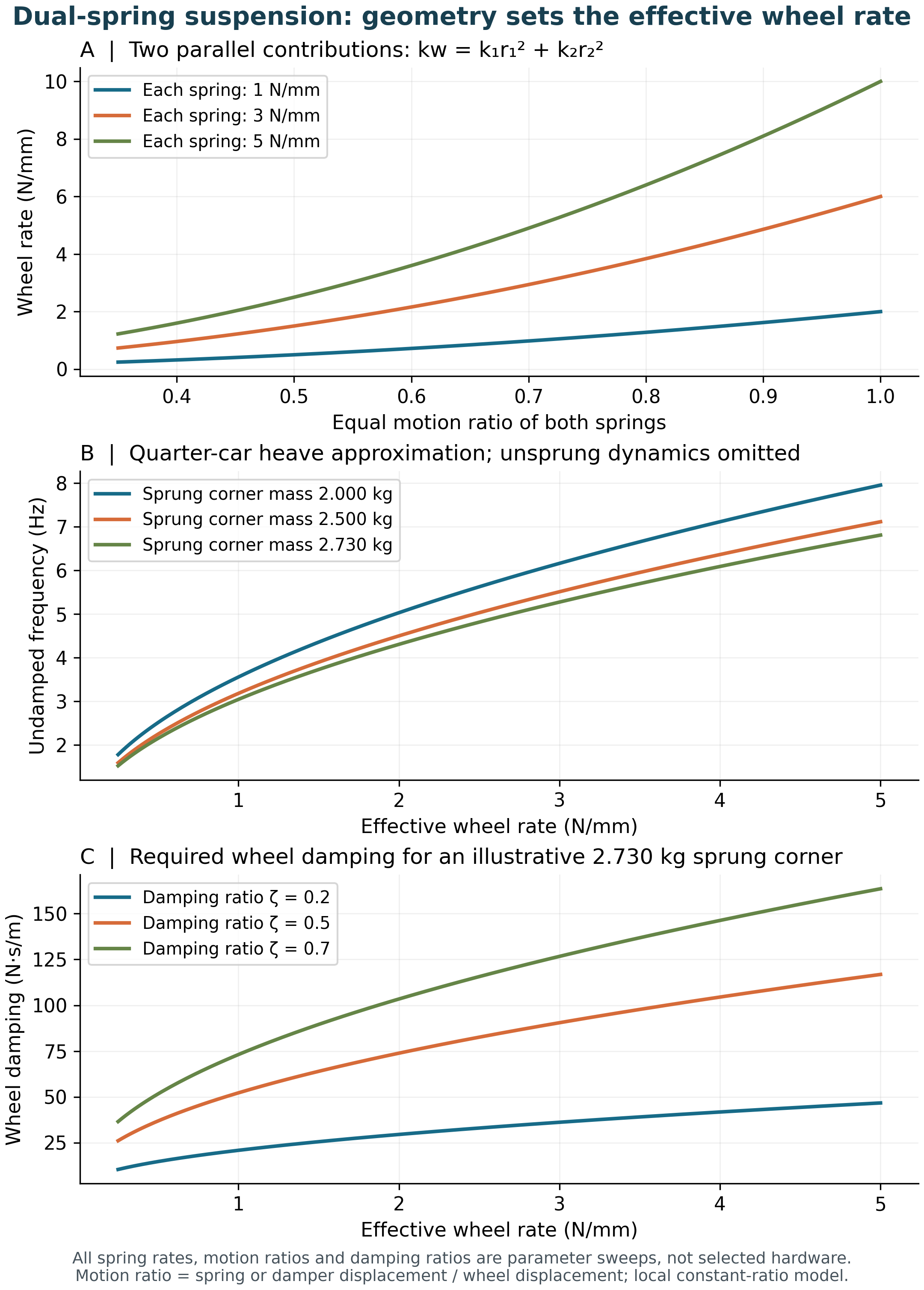}
\caption{Local suspension sensitivity for two springs and one damper per
wheel. Spring rates, motion ratios and sprung masses are illustrative
parameter sweeps; none is a measured response of the proposed hardware.}
\end{figure}

\hypertarget{power-distribution-protection-and-energy-budget}{%
\Needspace{4\baselineskip}
\section{6. Power Distribution, Protection, and Energy
Budget}\label{power-distribution-protection-and-energy-budget}}

\hypertarget{separate-propulsion-and-task-supplies}{%
\Needspace{3\baselineskip}
\subsection{6.1. Separate propulsion and task
supplies}\label{separate-propulsion-and-task-supplies}}

The propulsion pack is a 3S, 5,000 mAh lithium-polymer battery; the task
pack is a separate 3S, 2,200 mAh battery. Their nominal energies are

\[E_d=11.1(5.0)=55.50\ \mathrm{Wh},\qquad
E_t=11.1(2.2)=24.42\ \mathrm{Wh}.\]

The combined nominal energy is 79.92 Wh, but this total is not one
interchangeable energy reservoir. Unused drive energy cannot sustain the
task computer unless the architecture explicitly provides a suitable
transfer path, which it currently does not. Each battery needs its own
insulated enclosure, disconnect, restraint, and appropriately selected
protection. Full-charge voltage is 12.6 V; nominal voltage is not the
maximum regulator input condition.

The task pack supplies four separately regulated branches. The Pi branch
uses the EP-0225 power arrangement to provide the specified USB-C power
path for the Raspberry Pi 5 \cite{ep0225,raspberrypi5}. A 5 V UBEC supplies the
pan servo. Two DFR0379 regulators provide nominal 5.2 V rails for LiDAR
and tilt-servo branches. The Raspberry Pi Camera Module 3 connects
through the correct Pi 5 camera cable, while the display uses HDMI and
USB as appropriate. The selected hardware revision and connector
polarity must be checked before power is applied.

The PCA9685 breakout board has a logic supply and a separate external
servo-power arrangement. Its regulated positive servo rail must not
accidentally join the independently supplied pan and tilt rails. The
Pi-side I\textsuperscript{2}C logic and pull-ups remain at 3.3 V. An SN74AHCT125 powered
at 5 V can provide a TTL-compatible level interface for the pan control
signal \cite{ahct125}. The tilt servo's
acceptance of the intended logic level still requires verification; a
supply-voltage rating alone does not establish its input-high threshold.
All signal returns must follow the designed task return path.

\begin{figure}
\centering
\includegraphics[width=6.45in,height=\textheight]{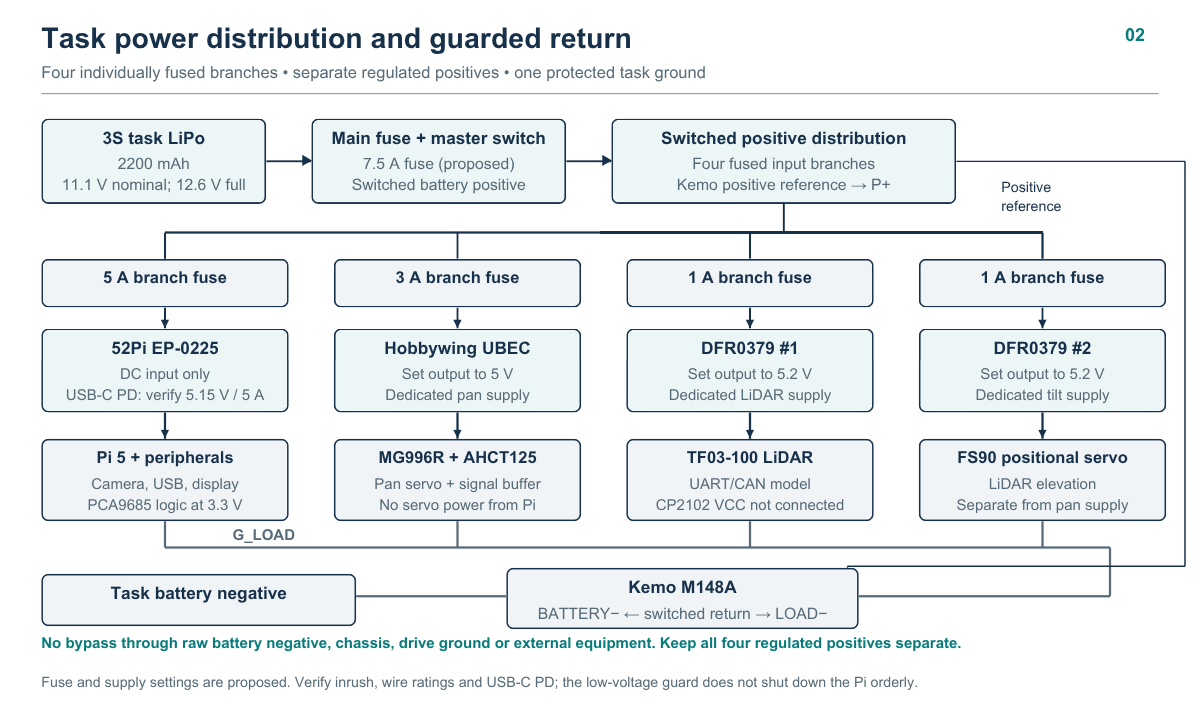}
\caption{Editable Canva task-power diagram. The separate positive rails
and the low-side battery guard are explicit. Branch labels describe
proposed hardware, not measured current or validated fuse coordination.}
\end{figure}

\hypertarget{low-side-guard-and-shutdown-behavior}{%
\Needspace{3\baselineskip}
\subsection{6.2. Low-side guard and shutdown
behavior}\label{low-side-guard-and-shutdown-behavior}}

The M148A guard switches the negative return path
\cite{kemo}. Its BATTERY− terminal belongs to
the raw task-battery negative, while all protected task loads return
through LOAD−. An unintended USB, chassis, instrument, or drive-system
ground connection can bypass that switching path. Bench equipment must
therefore be connected with awareness of the intended return topology.
The positive reference follows the switched positive distribution as
specified by the final wiring design.

A pack-level guard is not a cell-balancing system and does not replace
per-cell monitoring. Nor does sudden power removal constitute orderly
operating-system shutdown. The Pi must finish writing photographs and
event records before its supply is removed. A practical operating
procedure needs a verified voltage threshold, sufficient reserve for
shutdown, and testing under transient servo and USB loads. Because a
pack threshold does not directly reveal an individual weak cell, the
design should include the planned per-cell alarm and pre-run battery
checks.

Electrical integration tests should record the Pi input voltage,
regulator outputs, task-pack current, and relevant return-path voltages
while the pan servo starts, the camera captures, and the display is
active. A successful idle boot does not demonstrate immunity to servo
transients or radio-frequency interference. Active cooling should be
installed and its wiring secured; CPU throttling and converter heating
must be checked during the complete mission sequence.

\hypertarget{illustrative-current-and-runtime-envelopes}{%
\Needspace{3\baselineskip}
\subsection{6.3. Illustrative current and runtime
envelopes}\label{illustrative-current-and-runtime-envelopes}}

For an output branch power \(P_o\), battery voltage \(V_b\), and assumed
converter efficiency \(\eta\), input current is

\[I_b=\frac{P_o}{\eta V_b}.\]

Input current must be compared with an input fuse; output current must
be compared with the output rail and load wiring. Mixing those
quantities produces misleading protection calculations. The following
simultaneous-load example uses \(V_b=10.5\) V and \(\eta=0.85\). It is a
planning scenario, not an instrumented load profile.

\FloatBarrier
\begingroup\small
\begin{longtable}[]{@{}
  >{\raggedright\arraybackslash}p{(\columnwidth - 6\tabcolsep) * \real{0.2500}}
  >{\raggedleft\arraybackslash}p{(\columnwidth - 6\tabcolsep) * \real{0.2500}}
  >{\raggedleft\arraybackslash}p{(\columnwidth - 6\tabcolsep) * \real{0.2500}}
  >{\raggedleft\arraybackslash}p{(\columnwidth - 6\tabcolsep) * \real{0.2500}}@{}}
\caption{Illustrative task-branch load and battery-side current
budget.}\tabularnewline
\toprule\noalign{}
\begin{minipage}[b]{\linewidth}\raggedright
Task branch
\end{minipage} & \begin{minipage}[b]{\linewidth}\raggedleft
Illustrative output power (W)
\end{minipage} & \begin{minipage}[b]{\linewidth}\raggedleft
Calculated battery-side current (A)
\end{minipage} & \begin{minipage}[b]{\linewidth}\raggedleft
Proposed input fuse (A)
\end{minipage} \\
\midrule\noalign{}
\endfirsthead
\toprule\noalign{}
\begin{minipage}[b]{\linewidth}\raggedright
Task branch
\end{minipage} & \begin{minipage}[b]{\linewidth}\raggedleft
Illustrative output power (W)
\end{minipage} & \begin{minipage}[b]{\linewidth}\raggedleft
Calculated battery-side current (A)
\end{minipage} & \begin{minipage}[b]{\linewidth}\raggedleft
Proposed input fuse (A)
\end{minipage} \\
\midrule\noalign{}
\endhead
\bottomrule\noalign{}
\endlastfoot
Pi and USB branch & 25.75 & 2.885 & 5 \\
Pan-servo branch & 15.00 & 1.681 & 3 \\
LiDAR branch & 2.00 & 0.224 & 1 \\
Tilt-servo branch & 5.20 & 0.583 & 1 \\
Total & 47.95 & 5.373 & 7.5 main \\
\end{longtable}
\endgroup

The nominal difference between 7.5 A and 5.373 A is not a validated
margin against nuisance operation or cable damage. Fuse time--current
curves, wire ampacity, ambient temperature, regulator startup current,
servo stall current, and source fault current remain relevant.
Regulators can also limit or oscillate before a fuse operates. The
protection scheme must be assessed as a coordinated circuit.

Assuming 80\% usable nominal pack energy over a 12-minute mission gives
average battery-power budgets of 222 W for propulsion and 97.68 W for
tasks. With 85\% task conversion efficiency, the latter corresponds to
83.03 W of average output power. These are arithmetic energy ceilings
under assumed usable fractions. They do not demonstrate that either pack
can provide a particular current without sag, that the motors can
tolerate the heat, or that a regulator can deliver that output
continuously.

For a constant 100 W propulsion draw, the same model gives 26.64
minutes. A constant 25 W task output at 85\% efficiency gives 39.85
minutes. Real mission endurance requires a measured duty cycle,
temperature, cell balance, converter efficiency, and shutdown reserve.
The graphs retain these distinctions instead of treating an optimistic
capacity calculation as a completed runtime test.

\begin{figure}
\centering
\includegraphics[width=5.8in,height=\textheight]{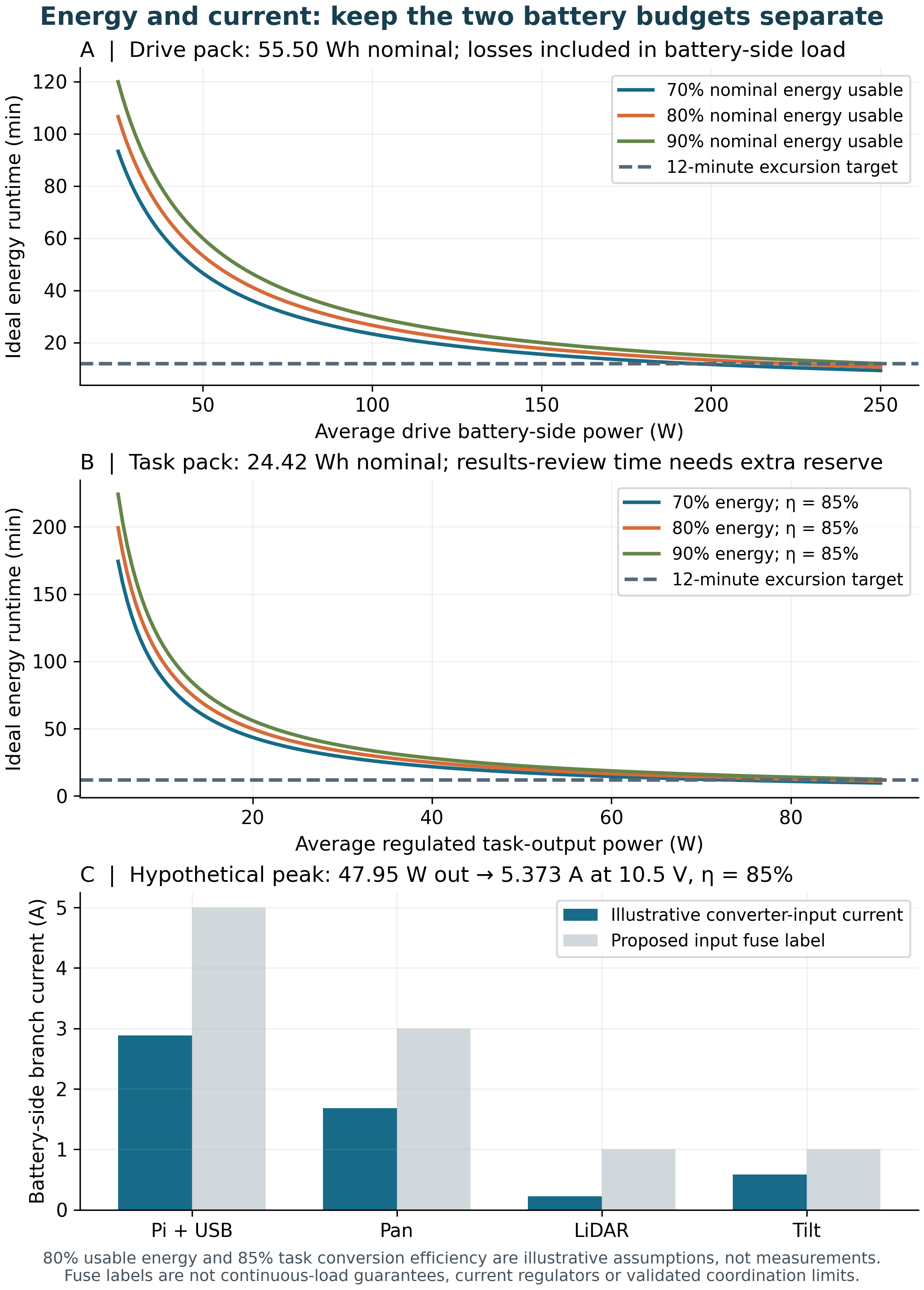}
\caption{Calculated energy and current scenarios for separate drive and
task packs. Assumed usable energy and conversion efficiency are shown;
the figures do not represent measured battery discharge or load traces.}
\end{figure}

\hypertarget{packet-commanded-imaging-and-lidar-ranging}{%
\Needspace{4\baselineskip}
\section{7. Packet-Commanded Imaging and LiDAR
Ranging}\label{packet-commanded-imaging-and-lidar-ranging}}

\hypertarget{task-sequence-and-receiving-architecture}{%
\Needspace{3\baselineskip}
\subsection{7.1. Task sequence and receiving
architecture}\label{task-sequence-and-receiving-architecture}}

The supplied handbook describes 4 ft × 4 ft white target boards carrying
a filled red circle, blue square, or green triangle. The red circle
diameter and blue square side are 12 inches; the equilateral green
triangle has a 12-inch height. Targets are placed within 50 ft of the
specified starting point. The task commands select pan motion,
photography, color filtering, and distance measurement
\cite{herc2027}. The software must therefore
preserve command order and state even when successive commands repeat a
letter.

The receiving chain uses a VHF antenna, an RTL-SDR V3 receiver,
demodulation, and Dire Wolf AX.25/APRS decoding
\cite{direwolf}. The operating description
specifies 144--148 MHz and 1200 baud; the final receiving frequency and
packet addressing must follow NASA's released configuration
\cite{herc2027}. The program should accept
validated decoded command tokens through an adapter, rather than
hard-code a guessed APRS payload format. A successful software decode in
a laboratory does not establish outdoor reception reliability in the
presence of motors, switching converters, USB devices, and nearby
transmitters.

\FloatBarrier
\begingroup\small
\begin{longtable}[]{@{}
  >{\raggedright\arraybackslash}p{(\columnwidth - 4\tabcolsep) * \real{0.3333}}
  >{\raggedright\arraybackslash}p{(\columnwidth - 4\tabcolsep) * \real{0.3333}}
  >{\raggedright\arraybackslash}p{(\columnwidth - 4\tabcolsep) * \real{0.3333}}@{}}
\caption{Received task-command semantics and execution
state.}\tabularnewline
\toprule\noalign{}
\begin{minipage}[b]{\linewidth}\raggedright
Token
\end{minipage} & \begin{minipage}[b]{\linewidth}\raggedright
Required operation represented in the executor
\end{minipage} & \begin{minipage}[b]{\linewidth}\raggedright
State consequence
\end{minipage} \\
\midrule\noalign{}
\endfirsthead
\toprule\noalign{}
\begin{minipage}[b]{\linewidth}\raggedright
Token
\end{minipage} & \begin{minipage}[b]{\linewidth}\raggedright
Required operation represented in the executor
\end{minipage} & \begin{minipage}[b]{\linewidth}\raggedright
State consequence
\end{minipage} \\
\midrule\noalign{}
\endhead
\bottomrule\noalign{}
\endlastfoot
AAA & Begin a task sequence & Open a new record after successful
initialization \\
B / C & Turn field of view 30° right / left & Update pan only after
successful motion \\
D & Capture and save a new photograph & Use current filter; attach
capture date/time \\
E / F & Select red / blue filter & Replace the prior filter selection \\
G & Select no filter & Restore normal-color output \\
H & Measure target distance & Record a fresh, valid result in feet \\
ZZZ & End the sequence & Finalize records for onboard review \\
\end{longtable}
\endgroup

The command letters and 30-second interval come from the task
specification; the prototype's internal 25-second service budget is an
engineering proposal, not a measured task latency. The reference timeout
applies after dispatch begins. Queue waiting and cancellation
acknowledgement are additional, so 25 seconds is not a hard
packet-arrival-to-result deadline. Reception should continue while a
command executes. A queue separates the receiver from motion, capture,
and ranging so that slow operations do not silently discard later
commands. Repeated B commands may legitimately request successive 30°
increments and must not be discarded merely because their payload
letters match.

\begin{figure}
\centering
\includegraphics[width=6.45in,height=\textheight]{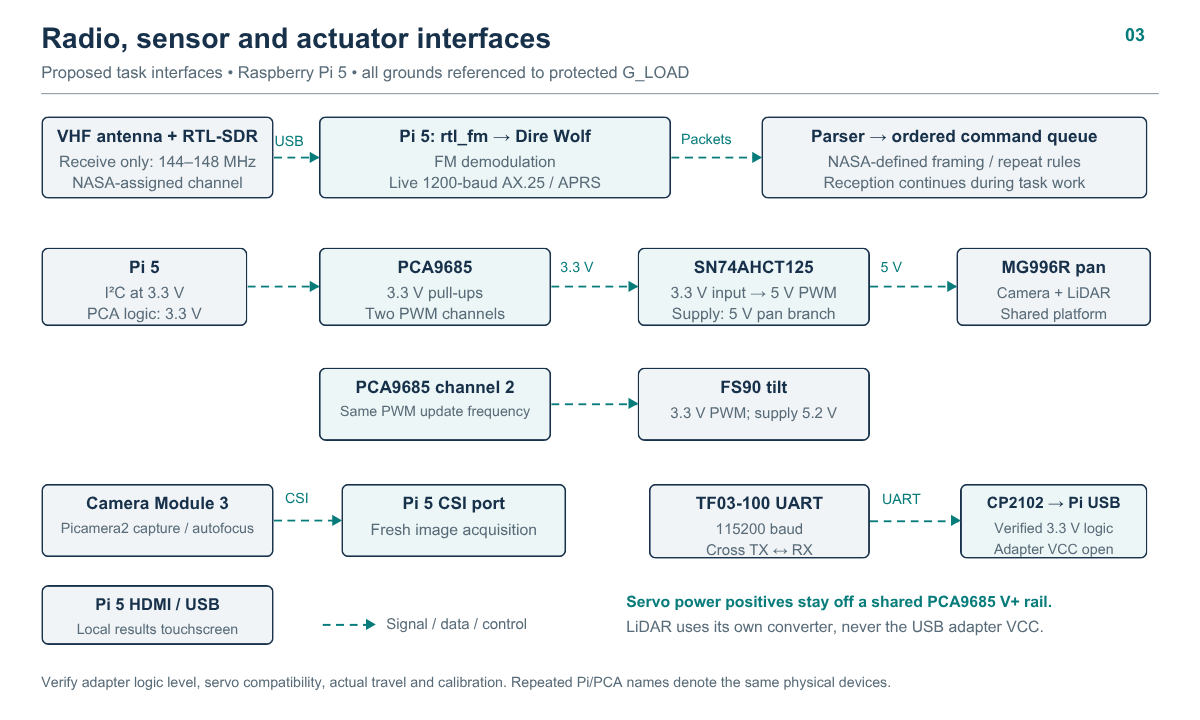}
\caption{Editable Canva task-interface diagram showing the receiving
chain, computer, camera, LiDAR, servo control, storage and display. This
is an integration specification; exact device revision and pin-level
checks remain necessary.}
\end{figure}

\hypertarget{fresh-photographs-and-filter-semantics}{%
\Needspace{3\baselineskip}
\subsection{7.2. Fresh photographs and filter
semantics}\label{fresh-photographs-and-filter-semantics}}

The camera is mounted on a bearing-supported pan assembly. The commanded
field-of-view change should correspond to measured physical camera
rotation. A hobby-servo pulse command alone does not establish angular
accuracy under backlash, cable load, or asymmetric mechanical stops. The
total pan range required by an unknown command sequence is not
established in the supplied description; consequently, a nominal 180°
servo must not be declared sufficient without resolving the permitted
sequence and calibrating usable travel.

For each D command, the program requests a new camera frame after the
selected orientation has settled. The frame's capture timestamp is
obtained from a correctly initialized clock, and the resulting image is
stored with the required date and time. The RTC cell preserves time
across power loss but does not itself guarantee that time was initially
set correctly. Unique file identifiers prevent a repeated photograph
command from overwriting earlier evidence.

Software red or blue filtering should be defined and demonstrated
consistently. One candidate channel-isolation transform is

\[I_{\mathrm{red}}=(R,0,0),\quad I_{\mathrm{blue}}=(0,0,B),\quad
I_{\mathrm{normal}}=(R,G,B).\]

This describes the proposed software interpretation; it is not an
assertion that the handbook mandates this exact pixel operation or that
software channel isolation is identical to a physical optical filter.
The chosen implementation should be confirmed against the task's
intended filter interpretation. A new filter replaces the previous one,
and ``no filter'' means ordinary color, not grayscale. The original
unfiltered frame is the appropriate input for the optional classifier,
because applying a red or blue output filter would destroy color
evidence needed for three-class target recognition.

\begin{figure}
\centering
\includegraphics[width=6.45in,height=\textheight]{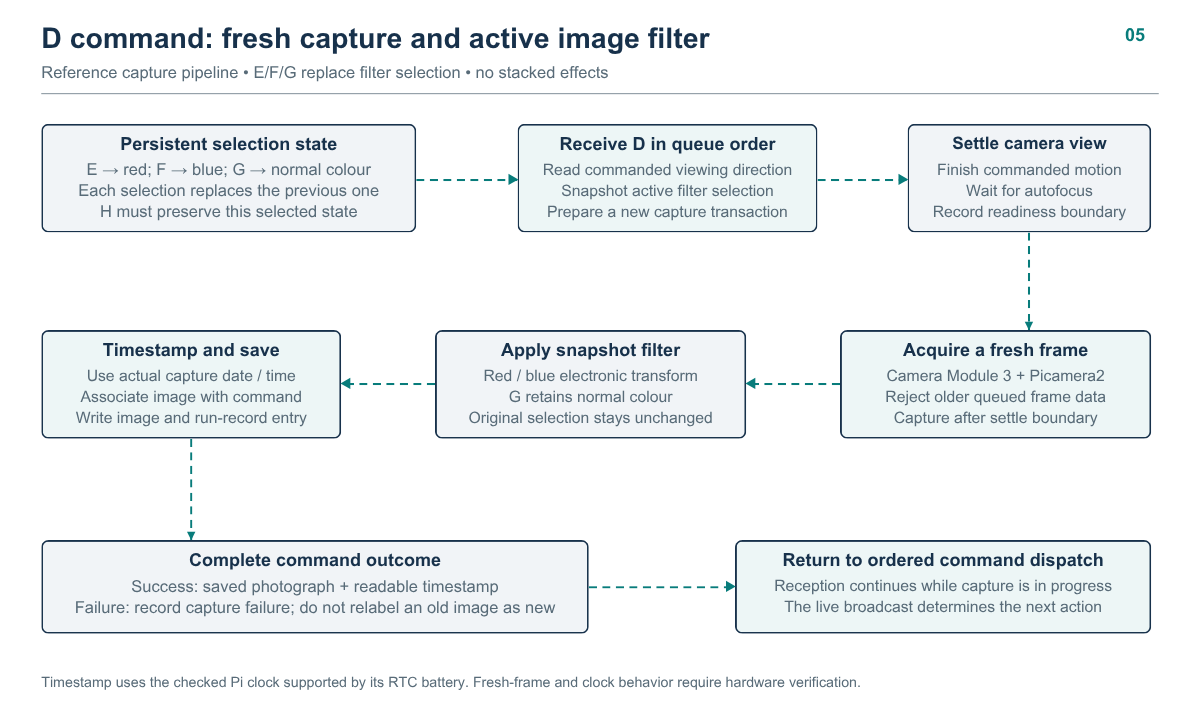}
\caption{Editable Canva capture pipeline. The command queue, current
filter, settled view, fresh frame, timestamp, storage and onboard review
are represented separately so that an image can be traced to its
command.}
\end{figure}

\hypertarget{pixel-sampling-and-camera-calibration}{%
\Needspace{3\baselineskip}
\subsection{7.3. Pixel sampling and camera
calibration}\label{pixel-sampling-and-camera-calibration}}

The Camera Module 3 Standard product brief specifies a 4608 × 2592
sensor image and a nominal horizontal field of view of 66°
\cite{camera3}. A full-frame pinhole
approximation gives focal length in pixels

\[f_x=\frac{N_x}{2\tan(\alpha_x/2)},\qquad p=\frac{f_xw}{Z},\]

where \(N_x=4608\), \(\alpha_x=66^{\circ}\), \(w\) is feature width, and \(Z\)
is optical-axis depth to a fronto-parallel feature, rather than general
Euclidean sensor-to-target range. For a 12-inch width at 50 ft,
\(p\approx70.96\) pixels. Scaling the same complete view to 1920 pixels
across gives approximately 29.57 pixels. This is a sampling estimate
rather than a recognition guarantee. Lens distortion, focus, exposure,
motion blur, compression, frame cropping, and the actual board pose
affect the image presented to an algorithm.

\FloatBarrier
\begingroup\small
\begin{longtable}[]{@{}
  >{\raggedleft\arraybackslash}p{(\columnwidth - 6\tabcolsep) * \real{0.2500}}
  >{\raggedleft\arraybackslash}p{(\columnwidth - 6\tabcolsep) * \real{0.2500}}
  >{\raggedleft\arraybackslash}p{(\columnwidth - 6\tabcolsep) * \real{0.2500}}
  >{\raggedleft\arraybackslash}p{(\columnwidth - 6\tabcolsep) * \real{0.2500}}@{}}
\caption{Ideal sampling of a 12-inch feature using the complete camera
view.}\tabularnewline
\toprule\noalign{}
\begin{minipage}[b]{\linewidth}\raggedleft
Range
\end{minipage} & \begin{minipage}[b]{\linewidth}\raggedleft
Angular width of a 12-inch feature
\end{minipage} & \begin{minipage}[b]{\linewidth}\raggedleft
Full 4608-pixel width estimate
\end{minipage} & \begin{minipage}[b]{\linewidth}\raggedleft
Same full view scaled to 1920 pixels
\end{minipage} \\
\midrule\noalign{}
\endfirsthead
\toprule\noalign{}
\begin{minipage}[b]{\linewidth}\raggedleft
Range
\end{minipage} & \begin{minipage}[b]{\linewidth}\raggedleft
Angular width of a 12-inch feature
\end{minipage} & \begin{minipage}[b]{\linewidth}\raggedleft
Full 4608-pixel width estimate
\end{minipage} & \begin{minipage}[b]{\linewidth}\raggedleft
Same full view scaled to 1920 pixels
\end{minipage} \\
\midrule\noalign{}
\endhead
\bottomrule\noalign{}
\endlastfoot
10 ft & 5.725° & 354.78 px & 147.83 px \\
25 ft & 2.292° & 141.91 px & 59.13 px \\
50 ft & 1.146° & 70.96 px & 29.57 px \\
\end{longtable}
\endgroup

The triangle's specified height is 12 inches; its equilateral side is
\(2h/\sqrt{3}\), not 12 inches. The table uses a generic 12-inch feature
dimension and should not be interpreted as assigning identical bounding
boxes to all shapes. Camera calibration should be performed at the
chosen capture mode and focus configuration with target views spanning
the usable image area
\cite{zhang2000calibration}. Reprojection
residuals quantify the fitted camera model; they do not independently
establish LiDAR alignment or task accuracy.

\begin{figure}
\centering
\includegraphics[width=6.45in,height=\textheight]{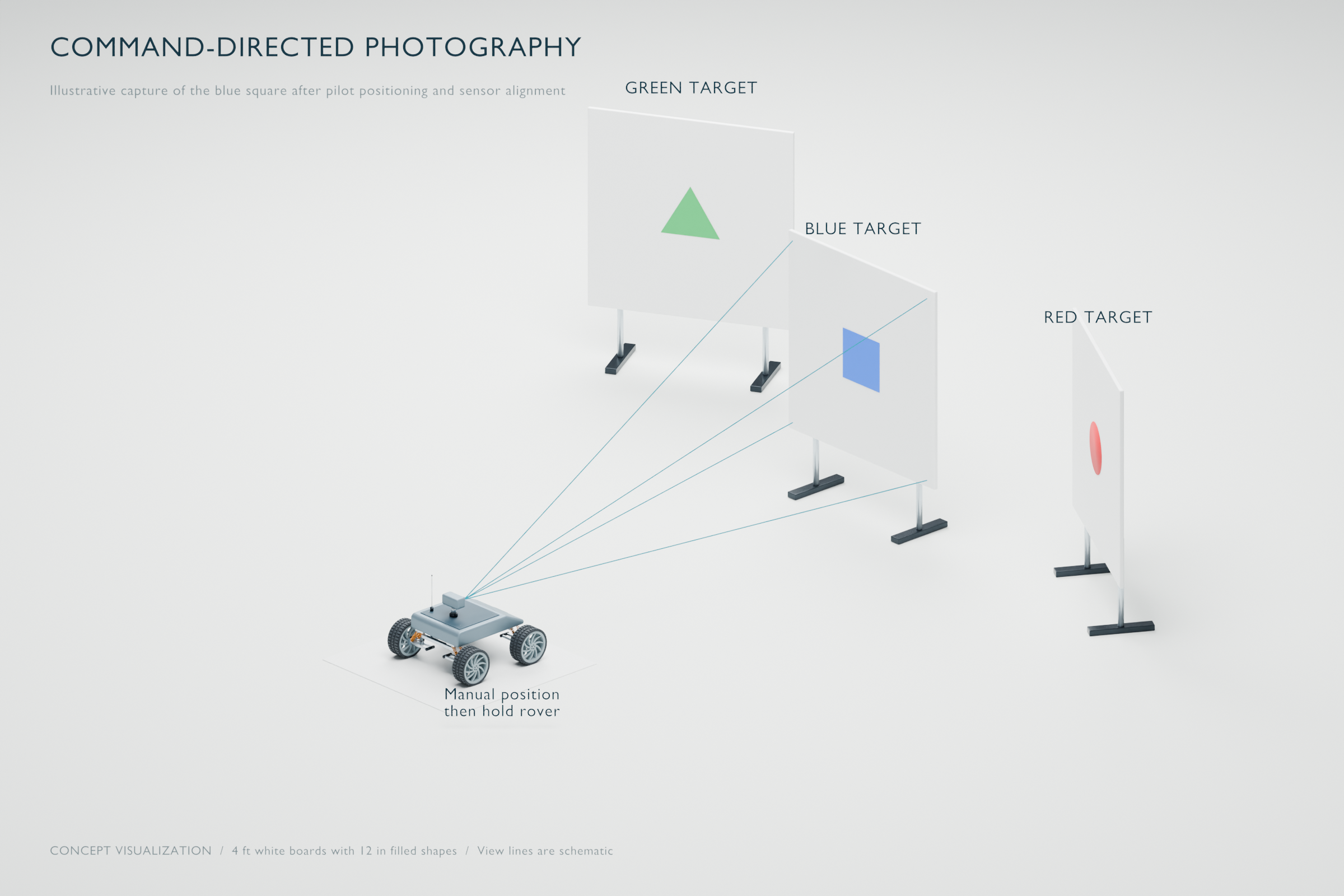}
\caption{Blender illustration of the photography task with the rover
stationary and illustrative target placements. The scene is a
visualization of the intended operation, not a rendered dataset used for
the classification experiment and not an observed competition run.}
\end{figure}

\hypertarget{single-beam-ranging-and-reference-geometry}{%
\Needspace{3\baselineskip}
\subsection{7.4. Single-beam ranging and reference
geometry}\label{single-beam-ranging-and-reference-geometry}}

The TF03-100 is a single-beam range sensor. Its listed field of view is
0.5° and its performance is conditional on target reflectivity and
environmental conditions \cite{tf03}.
Treating 0.5° as a full cone angle for a geometric illustration gives

\[b(R)=2R\tan(0.25^{\circ}).\]

Here \(R\) denotes the nominal perpendicular distance to the target
plane; the footprint calculation assumes the beam is normal to that
plane. For a pointing error \(\delta\), the ray's slant distance to the
same plane is \(R\sec\delta\). At 50 ft, this construction produces a
nominal width of about 133.0 mm. It is not a verified beam-profile
measurement. A pointing error \(\delta\) displaces the ray laterally by
\(R\tan\delta\). At 50 ft, 1° produces 266.0 mm and 3° produces 798.7
mm. The latter exceeds the half-width of a centered 4 ft board. Thus a
stable numerical range can still be associated with the wrong surface.

By contrast, if the ray still intersects the same flat plane, the
slant-minus-normal difference is \(R(\sec\delta-1)\), only 20.9 mm at 3°
and 50 ft. This comparison identifies board association and boresight
alignment as primary integration problems. Repeatedly averaging a
wrong-surface return does not fix them.

The ranging adapter should aim at the intended board, verify serial
framing and checksum, reject invalid or stale values, collect a bounded
set of fresh samples, and report failure explicitly if no valid
board-associated measurement is obtained. A CP2102 serial adapter must
use a compatible logic interface, and the LiDAR receives its specified
regulated supply separately. A physical calibration procedure must
determine camera intrinsics, the camera-to-LiDAR transform, servo-angle
mapping, mounting offsets, and the effect of sensor tilt. Multi-beam
calibration papers provide useful principles
\cite{geiger2012single,levinson2013online}, but their
observation models cannot simply be transplanted to a single beam.

Let \(\mathbf{s}\) be the sensor position relative to a chosen rover
reference, pending confirmation of the reporting datum, and
\(\mathbf{u}\) the calibrated unit beam direction. A measured range
\(\rho\) gives target position

\[\mathbf{q}=\mathbf{s}+\rho\mathbf{u}.\]

If the required reported quantity is Euclidean reference-to-target
distance, it is \(\|\mathbf{q}\|\); if the requirement is horizontal
distance, the horizontal projection is used. A fixed scalar subtraction
is generally wrong while the head pans. The correct reference definition
must be confirmed, and the human-powered division's ±3 ft allowance must
not be assigned to the RC task. Manufacturer accuracy figures are
likewise not NASA judging tolerances. At 50 ft, the TF03-100 datasheet's
1\% accuracy specification corresponds to ±0.5 ft under its stated
90\%-reflective-white-board conditions; this does not characterize the
complete mounted system.

\begin{figure}
\centering
\includegraphics[width=6.45in,height=\textheight]{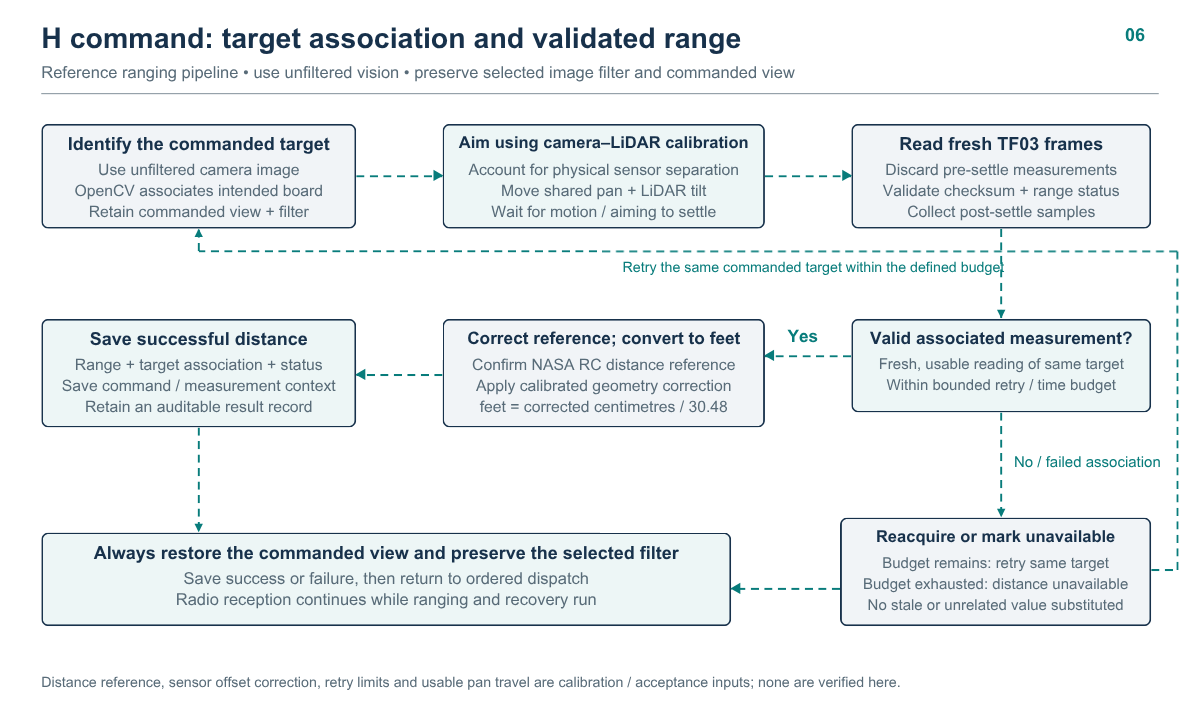}
\caption{Editable Canva ranging pipeline with target association,
calibration, frame validation, fresh sampling, bounded retry and
explicit unavailable-result handling. A plausible distance alone is
insufficient evidence of a correct target measurement.}
\end{figure}

\begin{figure}
\centering
\includegraphics[width=6.45in,height=\textheight]{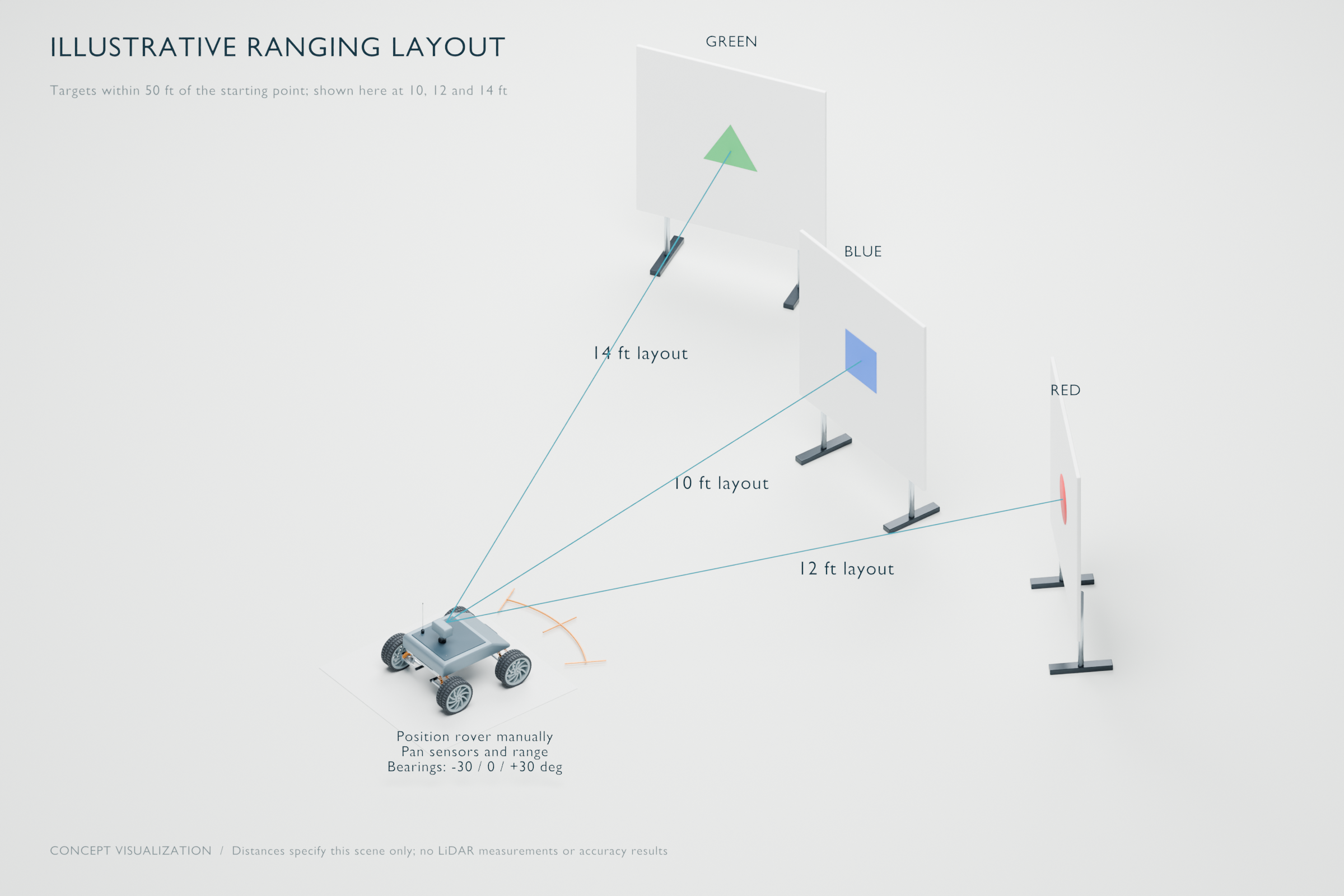}
\caption{Blender ranging illustration. The beam and annotated board
layout are schematic visualization elements; distances in the scene are
illustrative, not recorded TF03 measurements or a surveyed NASA course.}
\end{figure}

\begin{figure}
\centering
\includegraphics[width=5.8in,height=\textheight]{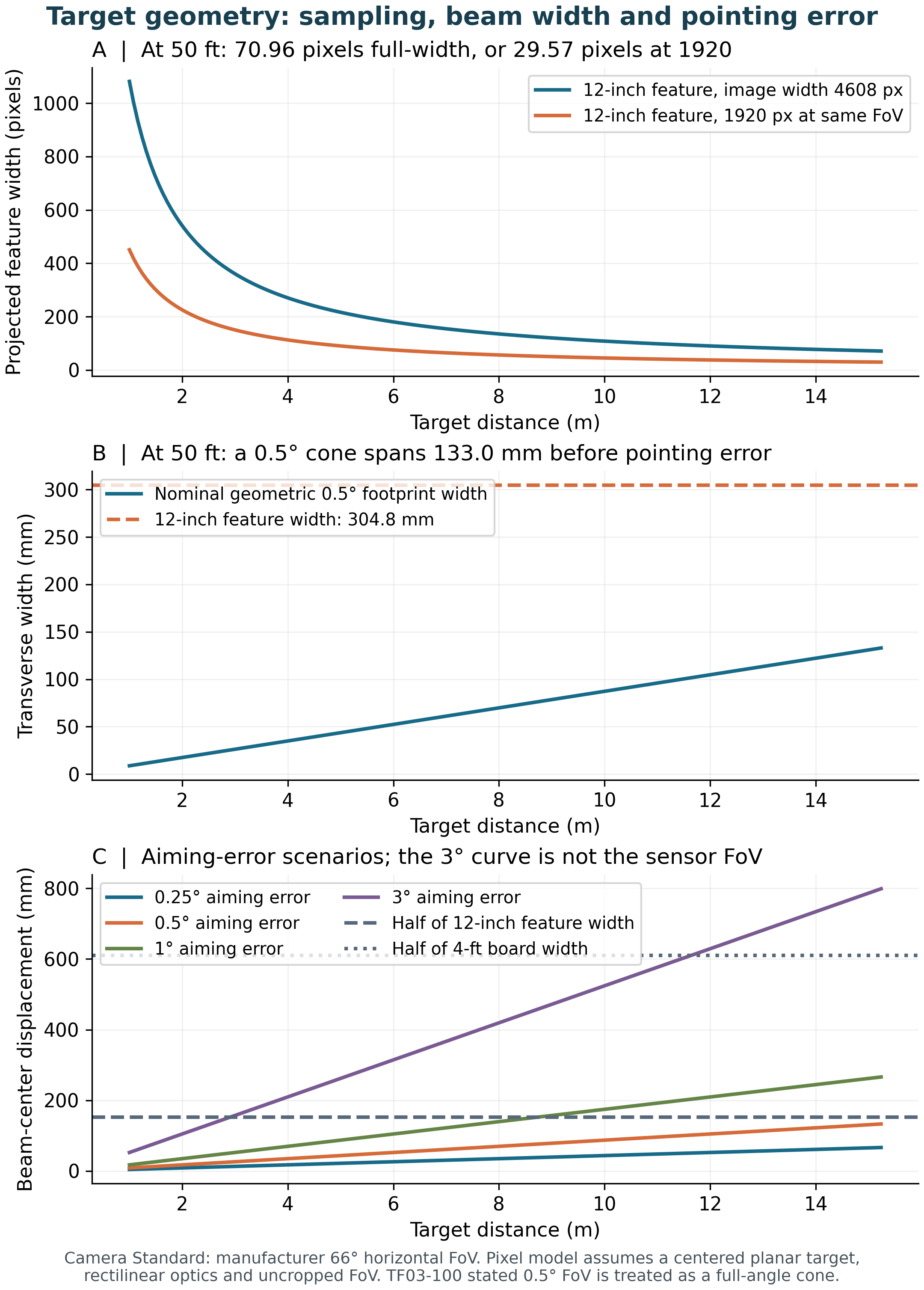}
\caption{Calculated camera sampling, nominal beam-width geometry and
aiming-error sensitivity. These are deterministic geometric calculations
with declared sensor specifications, not calibration residuals from the
rover.}
\end{figure}

\FloatBarrier

\hypertarget{reference-software-and-command-verification}{%
\Needspace{4\baselineskip}
\section{8. Reference Software and Command
Verification}\label{reference-software-and-command-verification}}

\hypertarget{state-model-and-implementation-boundary}{%
\Needspace{3\baselineskip}
\subsection{8.1. State model and implementation
boundary}\label{state-model-and-implementation-boundary}}

The accompanying Python reference executor uses WAITING, ACTIVE, and
FINALIZED states. A successful AAA record-opening operation establishes
ACTIVE. Filters are state variables; motion updates the commanded pan
only after its adapter succeeds. A serialized acceptance path prevents
overlapping commands from racing. A hardware error or timeout records an
explicit result and inhibits dependent motion, photography, and ranging
until the system has been deliberately recovered. Record finalization
remains possible so that partial evidence is retained. The reference
exposes a recovery-required flag but does not implement physical
recovery or an automatic flag-clearing procedure; these belong to the
hardware integration.

The executor receives already validated tokens. Radio packet addressing,
exact payload syntax, retransmission identity, and source and address
validation supported by the released protocol belong to the receiving
adapter and must follow NASA's released specification. The code does not
fabricate that wire format. Its event list is an in-memory trace, so the
deployed adapter must additionally persist the required records.
Cancellation of an asynchronous call cannot by itself guarantee that a
physical servo stops; the hardware interface must define bounded
operations, recovery, and safe behavior after communication loss.

The illustrative ±90° pan bounds in the reference code are test
parameters. They are not a conclusion that 180° travel satisfies all
possible commands. Out-of-range turns are rejected and logged, rather
than silently clamped or wrapped. This behavior exposes a design
deficiency instead of disguising it as successful execution. A complete
competition implementation must resolve mechanical range before
deployment.

\begin{figure}
\centering
\includegraphics[width=6.45in,height=\textheight]{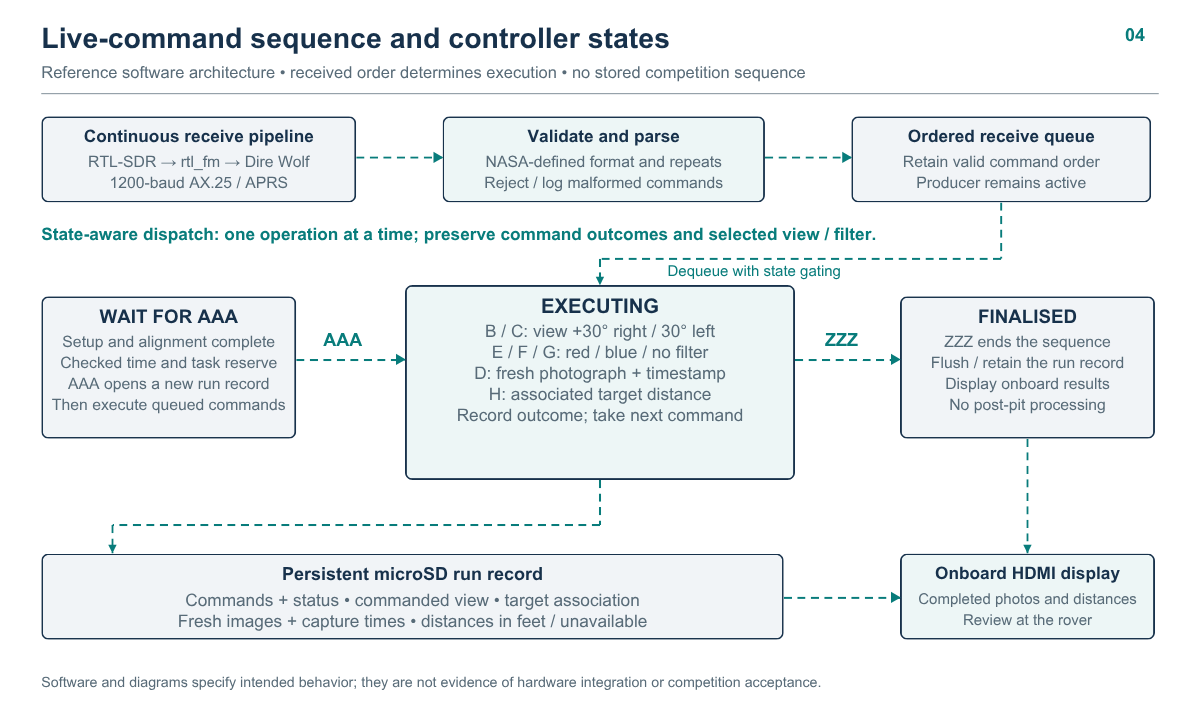}
\caption{Editable Canva command-state and fault-flow diagram. Reception
and execution are decoupled, and invalid, out-of-range or failed
operations remain visible in the record.}
\end{figure}

The following extract illustrates state replacement, bounded motion, and
explicit unit conversion. It is explanatory code from the
hardware-independent design, not a complete radio or motor driver.

\begin{minipage}{\linewidth}
\begin{Shaded}
\begin{Highlighting}[]
\ControlFlowTok{if}\NormalTok{ token }\KeywordTok{in}\NormalTok{ \{}\StringTok{"E"}\NormalTok{, }\StringTok{"F"}\NormalTok{, }\StringTok{"G"}\NormalTok{\}:}
\NormalTok{    image\_filter }\OperatorTok{=}\NormalTok{ \{}\StringTok{"E"}\NormalTok{: }\StringTok{"red"}\NormalTok{, }\StringTok{"F"}\NormalTok{: }\StringTok{"blue"}\NormalTok{, }\StringTok{"G"}\NormalTok{: }\StringTok{"normal"}\NormalTok{\}[token]}
\ControlFlowTok{elif}\NormalTok{ token }\KeywordTok{in}\NormalTok{ \{}\StringTok{"B"}\NormalTok{, }\StringTok{"C"}\NormalTok{\}:}
\NormalTok{    target }\OperatorTok{=}\NormalTok{ pan\_deg }\OperatorTok{+}\NormalTok{ (}\FloatTok{30.0} \ControlFlowTok{if}\NormalTok{ token }\OperatorTok{==} \StringTok{"B"} \ControlFlowTok{else} \OperatorTok{{-}}\FloatTok{30.0}\NormalTok{)}
    \ControlFlowTok{if} \KeywordTok{not}\NormalTok{ pan\_min }\OperatorTok{\textless{}=}\NormalTok{ target }\OperatorTok{\textless{}=}\NormalTok{ pan\_max:}
        \ControlFlowTok{return}\NormalTok{ \{}\StringTok{"status"}\NormalTok{: }\StringTok{"rejected"}\NormalTok{, }\StringTok{"reason"}\NormalTok{: }\StringTok{"pan\_range"}\NormalTok{\}}
    \ControlFlowTok{await}\NormalTok{ hardware.pan\_to(target)}
\NormalTok{    pan\_deg }\OperatorTok{=}\NormalTok{ target}
\ControlFlowTok{elif}\NormalTok{ token }\OperatorTok{==} \StringTok{"H"}\NormalTok{:}
\NormalTok{    range\_cm }\OperatorTok{=} \ControlFlowTok{await}\NormalTok{ hardware.measure\_target(pan\_deg)}
\NormalTok{    valid }\OperatorTok{=}\NormalTok{ (}\BuiltInTok{isinstance}\NormalTok{(range\_cm, (}\BuiltInTok{int}\NormalTok{, }\BuiltInTok{float}\NormalTok{))}
             \KeywordTok{and} \KeywordTok{not} \BuiltInTok{isinstance}\NormalTok{(range\_cm, }\BuiltInTok{bool}\NormalTok{)}
             \KeywordTok{and}\NormalTok{ math.isfinite(range\_cm) }\KeywordTok{and}\NormalTok{ range\_cm }\OperatorTok{\textgreater{}} \DecValTok{0}\NormalTok{)}
    \ControlFlowTok{if} \KeywordTok{not}\NormalTok{ valid:}
        \ControlFlowTok{return}\NormalTok{ \{}\StringTok{"status"}\NormalTok{: }\StringTok{"distance\_unavailable"}\NormalTok{\}}
    \ControlFlowTok{return}\NormalTok{ \{}\StringTok{"status"}\NormalTok{: }\StringTok{"range\_saved"}\NormalTok{, }\StringTok{"distance\_ft"}\NormalTok{: range\_cm }\OperatorTok{/} \FloatTok{30.48}\NormalTok{\}}
\end{Highlighting}
\end{Shaded}
\end{minipage}\par

The conversion factor is exact because 1 ft = 30.48 cm. A range adapter
must return a correctly referenced value in the declared unit; the
executor cannot repair a wrong datum or mixed millimeter/centimeter
input merely by dividing. Successful execution should also preserve the
requested camera pan and filter across any temporary LiDAR targeting
motion. The current software checks logical state preservation; physical
restoration needs hardware measurement.

\hypertarget{executed-checks-and-deployment-tests}{%
\Needspace{3\baselineskip}
\subsection{8.2. Executed checks and deployment
tests}\label{executed-checks-and-deployment-tests}}

The verification script executes 30 focused checks with mocked hardware.
Cases include commands outside an active sequence, repeated 30° turns,
filter replacement, distinct photo requests, valid and invalid ranges,
unknown tokens, sequence finalization, queue order, out-of-range motion,
timeout inhibition, record-opening failure, concurrent acceptance, and
cancellation. Boolean values are rejected as distances even though
Python treats booleans as integer subclasses. A photo-adapter result
cannot silently overwrite the executor's own status.

These checks are software unit-level evidence. They do not exercise an
RF receiver, physical servo, camera exposure, actual storage device, or
LiDAR target. They nevertheless provide a reproducible foundation for
integration because they define failure behavior before physical tests.
The next stage is to replace mocks with device adapters while preserving
the same event schema and to replay complete sequences under controlled
fault injection.

A useful commissioning sequence starts with the rover mechanically
supported, then proceeds to stationary task operation, and only later to
combined drive and task tests. Inject malformed serial frames, remove
the LiDAR briefly, force an image-write error, and test recovery after a
supply interruption. Confirm that the onboard display exposes missing
data rather than substituting old images or zero-valued ranges. The
final acceptance demonstration must show fresh results after an unknown
sequence, with no external editing or return to the pit.

\hypertarget{synthetic-evaluation-of-learning-assisted-target-recognition}{%
\Needspace{4\baselineskip}
\section{9. Synthetic Evaluation of Learning-Assisted Target
Recognition}\label{synthetic-evaluation-of-learning-assisted-target-recognition}}

\hypertarget{dataset-and-experimental-separation}{%
\Needspace{3\baselineskip}
\subsection{9.1. Dataset and experimental
separation}\label{dataset-and-experimental-separation}}

The optional recognition extension was evaluated on an executed
synthetic benchmark because no physical camera recordings or target
trials were supplied. The task is four-way classification of a supplied
board crop: red circle, blue square, green triangle, or no target. It
does not test full-frame board detection, autonomous steering, or
physical Raspberry Pi integration. A procedural image generator creates
64 × 64 RGB crops containing a projectively transformed light board,
generated background, and a marker or negative example. It renders at
double resolution before downsampling, blur, and additive noise.

There are 2,400 images in 600 scene groups. Each group contains all four
labels with shared nuisance parameters, and each entire group belongs to
one split. This prevents the same pose, background, and lighting
configuration from leaking across training and evaluation. The negative
class contains blank boards or one of six incorrect color--shape
combinations. Those are deliberately difficult negatives, not additional
NASA target classes. Positive shapes have matched sampled area before
projection, reducing an avoidable area cue; the generated geometry is
not a metrically exact reproduction of all handbook board dimensions.

\FloatBarrier
\begingroup\small
\begin{longtable}[]{@{}
  >{\raggedright\arraybackslash}p{(\columnwidth - 6\tabcolsep) * \real{0.2500}}
  >{\raggedleft\arraybackslash}p{(\columnwidth - 6\tabcolsep) * \real{0.2500}}
  >{\raggedleft\arraybackslash}p{(\columnwidth - 6\tabcolsep) * \real{0.2500}}
  >{\raggedright\arraybackslash}p{(\columnwidth - 6\tabcolsep) * \real{0.2500}}@{}}
\caption{Scene-disjoint synthetic dataset allocation.}\tabularnewline
\toprule\noalign{}
\begin{minipage}[b]{\linewidth}\raggedright
Split
\end{minipage} & \begin{minipage}[b]{\linewidth}\raggedleft
Scene groups
\end{minipage} & \begin{minipage}[b]{\linewidth}\raggedleft
Images
\end{minipage} & \begin{minipage}[b]{\linewidth}\raggedright
Purpose
\end{minipage} \\
\midrule\noalign{}
\endfirsthead
\toprule\noalign{}
\begin{minipage}[b]{\linewidth}\raggedright
Split
\end{minipage} & \begin{minipage}[b]{\linewidth}\raggedleft
Scene groups
\end{minipage} & \begin{minipage}[b]{\linewidth}\raggedleft
Images
\end{minipage} & \begin{minipage}[b]{\linewidth}\raggedright
Purpose
\end{minipage} \\
\midrule\noalign{}
\endhead
\bottomrule\noalign{}
\endlastfoot
Training & 300 & 1,200 & Fit feature standardization and SVM
parameters \\
Validation & 100 & 400 & Select declared hyperparameters \\
Nominal test & 100 & 400 & Evaluate unseen scenes from nominal generator
ranges \\
Shifted test & 100 & 400 & Evaluate altered backgrounds and stronger
corruptions \\
\end{longtable}
\endgroup

Each test set has 100 examples per label. The generator varies board
size, rotation, perspective, marker location, color gain, exposure,
illumination gradients, blur, and noise. The shifted set uses held-out
checkerboard and diagonal-stripe background families, broader rotations,
stronger color changes, and increased blur and noise. All split seed
ranges are disjoint and all final image hashes are unique. This design
follows the motivation of appearance randomization
\cite{tobin2017domain,tremblay2018synthetic}, but the altered
test generator is not a real-world domain.

\FloatBarrier
\begingroup\small
\begin{longtable}[]{@{}
  >{\raggedright\arraybackslash}p{(\columnwidth - 4\tabcolsep) * \real{0.3333}}
  >{\raggedright\arraybackslash}p{(\columnwidth - 4\tabcolsep) * \real{0.3333}}
  >{\raggedright\arraybackslash}p{(\columnwidth - 4\tabcolsep) * \real{0.3333}}@{}}
\caption{Selected appearance ranges in the nominal and shifted
generators.}\tabularnewline
\toprule\noalign{}
\begin{minipage}[b]{\linewidth}\raggedright
Nuisance parameter
\end{minipage} & \begin{minipage}[b]{\linewidth}\raggedright
Nominal range
\end{minipage} & \begin{minipage}[b]{\linewidth}\raggedright
Shifted range
\end{minipage} \\
\midrule\noalign{}
\endfirsthead
\toprule\noalign{}
\begin{minipage}[b]{\linewidth}\raggedright
Nuisance parameter
\end{minipage} & \begin{minipage}[b]{\linewidth}\raggedright
Nominal range
\end{minipage} & \begin{minipage}[b]{\linewidth}\raggedright
Shifted range
\end{minipage} \\
\midrule\noalign{}
\endhead
\bottomrule\noalign{}
\endlastfoot
Board width / image width & 0.72--0.88 & 0.64--0.83 \\
Rotation & −30° to +30° & −55° to +55° \\
Per-corner projective jitter & ±0.045 image widths & ±0.10 image
widths \\
Exposure multiplier & 0.72--1.20 & 0.38--1.40 \\
Per-channel gain & 0.85--1.15 & 0.58--1.40 \\
Blur standard deviation & 0--0.65 output pixels & 0.65--1.65 output
pixels \\
Noise standard deviation & 0--4 gray levels & 3--12 gray levels \\
\end{longtable}
\endgroup

Because several nuisance factors change together, the experiment cannot
attribute performance loss to one corruption. The source package
includes the exact sampled parameters and RGB hashes for every example,
allowing the evaluation to be reproduced or separated into later
controlled ablations.

\begin{figure}
\centering
\includegraphics[width=6.45in,height=\textheight]{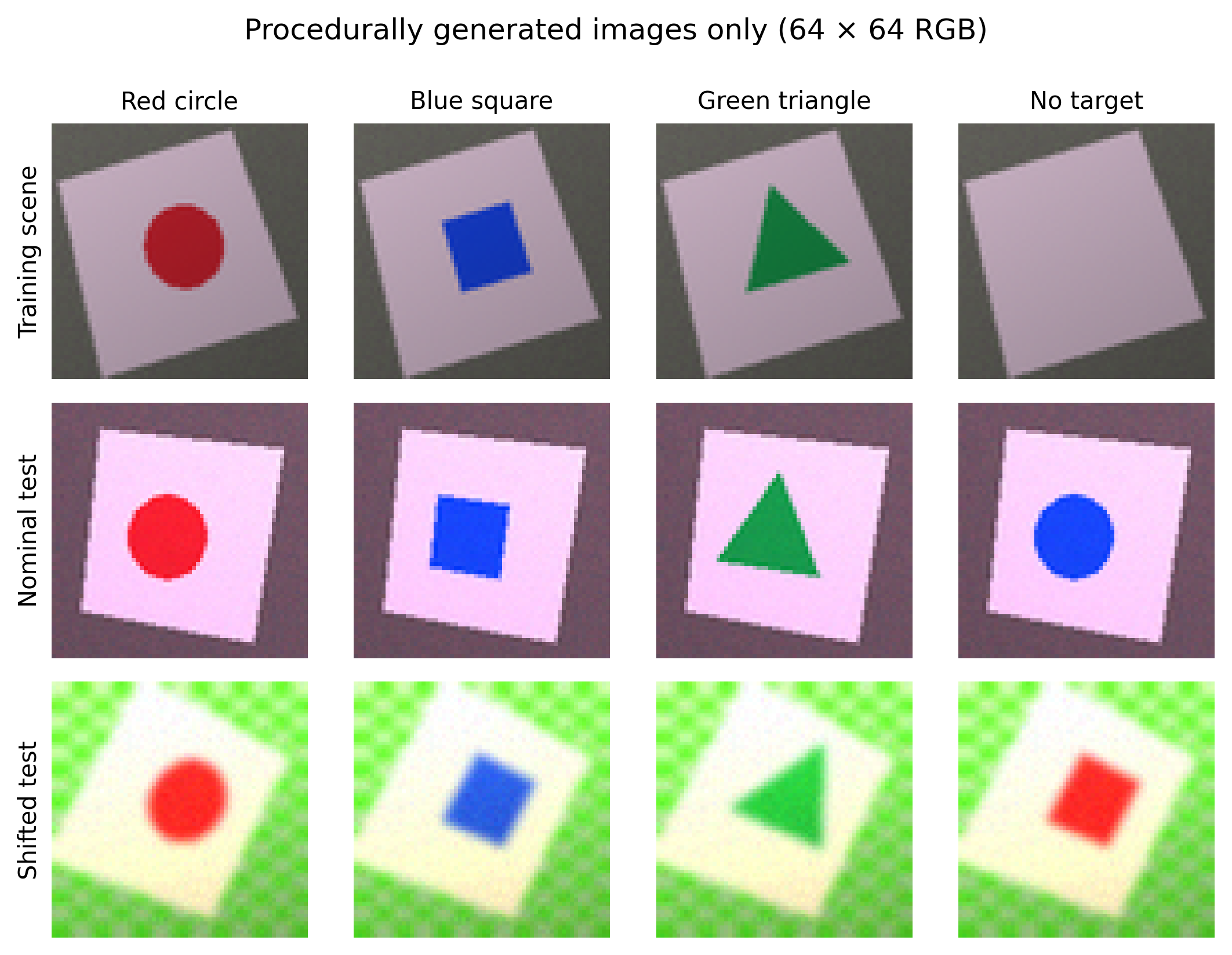}
\caption{Examples from the executed synthetic dataset. Nominal and
shifted board crops include target markers and no-target negatives.
These are procedurally generated images, not photographs from the rover
camera or Blender task scenes.}
\end{figure}

\hypertarget{features-classifiers-and-model-selection}{%
\Needspace{3\baselineskip}
\subsection{9.2. Features, classifiers, and model
selection}\label{features-classifiers-and-model-selection}}

Three compact methods are compared. The HSV-and-geometry baseline
thresholds colored regions, performs morphological cleanup, identifies
connected components, and evaluates convex-hull circularity,
oriented-box fill, and a heuristic solidity measure. Validation selects
three geometric cutoffs from 27 declared combinations. This method is
transparent and uses task-specific shape knowledge, but its thresholds
can be affected by lighting, occlusion, and low pixel resolution.

The first learned method uses a normalized 18 × 3 hue--saturation
histogram and eight value bins, giving 62 features. The second appends
1,764 grayscale HOG features, giving 1,826 total inputs. The HOG variant
uses nine unsigned orientation bins, 8 × 8-pixel cells, overlapping 2 ×
2-cell blocks, and L2-Hys normalization. Orientation votes are
interpolated; spatial vote interpolation is omitted. It is therefore a
documented HOG variant rather than an exact reimplementation of Dalal
and Triggs \cite{hog2005}.

Both learned methods use linear support-vector classifiers
\cite{svm1995}. For each one-versus-rest
binary problem, the squared-hinge formulation has the form

\[\min_{\mathbf{w},b}\ \frac{1}{2}\left(\|\mathbf{w}\|^2+b^2\right)+
C\sum_i\max\bigl(0,1-y_i(\mathbf{w}^{\mathsf T}\tilde{\mathbf{x}}_i+b)\bigr)^2,\]

where \(\tilde{\mathbf{x}}_i\) is standardized using training-set
statistics and \(y_i\in\{-1,+1\}\). With \texttt{fit\_intercept=True}
and the default \texttt{intercept\_scaling=1}, \texttt{LinearSVC}
penalizes the intercept through a constant synthetic feature. Multiclass
prediction selects the largest class decision value. Decision values and
gaps between them are not calibrated probabilities.

For each SVM, validation macro-F1 selects \(C\in\{0.0001,0.001,0.01\}\),
with ties choosing the smallest value. The final bounded grid follows a
computational pilot in which larger-C fits were slow; no HOG test score
informed that change. Interrupted pilots emitted color-only evaluation
metrics; the grid change was based on solver runtime. The experiment is
exploratory rather than preregistered. Models are fitted on training
examples only and are not refitted on validation data. The declared
procedure can be expressed compactly as follows.

\begin{minipage}{\linewidth}
\begin{Shaded}
\begin{Highlighting}[]
\NormalTok{best }\OperatorTok{=} \VariableTok{None}
\ControlFlowTok{for}\NormalTok{ C }\KeywordTok{in}\NormalTok{ (}\FloatTok{0.0001}\NormalTok{, }\FloatTok{0.001}\NormalTok{, }\FloatTok{0.01}\NormalTok{):}
\NormalTok{    model }\OperatorTok{=}\NormalTok{ make\_pipeline(}
\NormalTok{        StandardScaler(),}
\NormalTok{        LinearSVC(C}\OperatorTok{=}\NormalTok{C, dual}\OperatorTok{=}\VariableTok{False}\NormalTok{, max\_iter}\OperatorTok{=}\DecValTok{2000}\NormalTok{,}
\NormalTok{                  random\_state}\OperatorTok{=}\DecValTok{1703}\NormalTok{))}
\NormalTok{    model.fit(X\_train, y\_train)}
\NormalTok{    score }\OperatorTok{=}\NormalTok{ f1\_score(y\_valid, model.predict(X\_valid), average}\OperatorTok{=}\StringTok{"macro"}\NormalTok{)}
    \ControlFlowTok{if}\NormalTok{ best }\KeywordTok{is} \VariableTok{None} \KeywordTok{or}\NormalTok{ score }\OperatorTok{\textgreater{}}\NormalTok{ best[}\DecValTok{0}\NormalTok{]:}
\NormalTok{        best }\OperatorTok{=}\NormalTok{ (score, C, model)  }\CommentTok{\# ascending C resolves ties}
\NormalTok{selected\_model }\OperatorTok{=}\NormalTok{ best[}\DecValTok{2}\NormalTok{]}
\NormalTok{nominal\_predictions }\OperatorTok{=}\NormalTok{ selected\_model.predict(X\_test)}
\NormalTok{shifted\_predictions }\OperatorTok{=}\NormalTok{ selected\_model.predict(X\_shift)}
\end{Highlighting}
\end{Shaded}
\end{minipage}\par

The displayed listing summarizes model selection; the complete
generator, HOG implementation, geometric classifier, metrics, and
reproducibility controls are in the accompanying source. For each class
\(k\),

\[F_{1,k}=\frac{2TP_k}{2TP_k+FP_k+FN_k},\qquad
F_{1,\mathrm{macro}}=\frac{1}{4}\sum_{k=1}^{4}F_{1,k}.\]

Accuracy and macro-F1 are recomputed from saved predictions. The 95\%
percentile intervals use 1,000 bootstrap resamples of complete
four-image scene groups. These intervals represent scene-sampling
variation conditional on the fitted models; they do not include training
randomness, alternate generator choices, or physical-camera uncertainty.

\hypertarget{results-and-the-value-of-a-negative-comparison}{%
\Needspace{3\baselineskip}
\subsection{9.3. Results and the value of a negative
comparison}\label{results-and-the-value-of-a-negative-comparison}}

\FloatBarrier
\begingroup\small
\begin{longtable}[]{@{}
  >{\raggedright\arraybackslash}p{(\columnwidth - 8\tabcolsep) * \real{0.2000}}
  >{\raggedleft\arraybackslash}p{(\columnwidth - 8\tabcolsep) * \real{0.2000}}
  >{\raggedleft\arraybackslash}p{(\columnwidth - 8\tabcolsep) * \real{0.2000}}
  >{\raggedleft\arraybackslash}p{(\columnwidth - 8\tabcolsep) * \real{0.2000}}
  >{\raggedleft\arraybackslash}p{(\columnwidth - 8\tabcolsep) * \real{0.2000}}@{}}
\caption{Executed classification results on the two synthetic test
sets.}\tabularnewline
\toprule\noalign{}
\begin{minipage}[b]{\linewidth}\raggedright
Method
\end{minipage} & \begin{minipage}[b]{\linewidth}\raggedleft
Nominal accuracy
\end{minipage} & \begin{minipage}[b]{\linewidth}\raggedleft
Nominal macro-F1
\end{minipage} & \begin{minipage}[b]{\linewidth}\raggedleft
Shifted accuracy
\end{minipage} & \begin{minipage}[b]{\linewidth}\raggedleft
Shifted macro-F1
\end{minipage} \\
\midrule\noalign{}
\endfirsthead
\toprule\noalign{}
\begin{minipage}[b]{\linewidth}\raggedright
Method
\end{minipage} & \begin{minipage}[b]{\linewidth}\raggedleft
Nominal accuracy
\end{minipage} & \begin{minipage}[b]{\linewidth}\raggedleft
Nominal macro-F1
\end{minipage} & \begin{minipage}[b]{\linewidth}\raggedleft
Shifted accuracy
\end{minipage} & \begin{minipage}[b]{\linewidth}\raggedleft
Shifted macro-F1
\end{minipage} \\
\midrule\noalign{}
\endhead
\bottomrule\noalign{}
\endlastfoot
HSV + geometry & 95.00\% & 0.9512 & 71.75\% & 0.7343 \\
HSV histogram + SVM & 82.00\% & 0.7892 & 61.75\% & 0.5838 \\
HOG + HSV + SVM & 82.00\% & 0.7890 & 48.50\% & 0.4784 \\
\end{longtable}
\endgroup

The explicit geometric baseline performs best on both test sets. Adding
HOG provides no nominal-accuracy improvement over the color histogram
and reduces shifted accuracy by 13.25 percentage points. Both learned
models correctly reject only 34 of 100 nominal no-target examples. A
report restricted to positive-marker accuracy would therefore conceal a
significant failure mode. The selected C values are 0.001 for the color
model and 0.0001 for the combined model.

The nominal accuracy interval is 91.00--98.25\% for the geometric
baseline and 79.50--84.75\% for the HOG-plus-color model. Under the
shifted generator, the corresponding intervals are 65.50--78.00\% and
43.49--52.75\%. These are group-bootstrap intervals, not field success
bounds. The result does not establish that geometry will always
outperform learning; it shows that adding this feature representation
and linear classifier did not improve this declared benchmark.

\begin{figure}
\centering
\includegraphics[width=6.45in,height=\textheight]{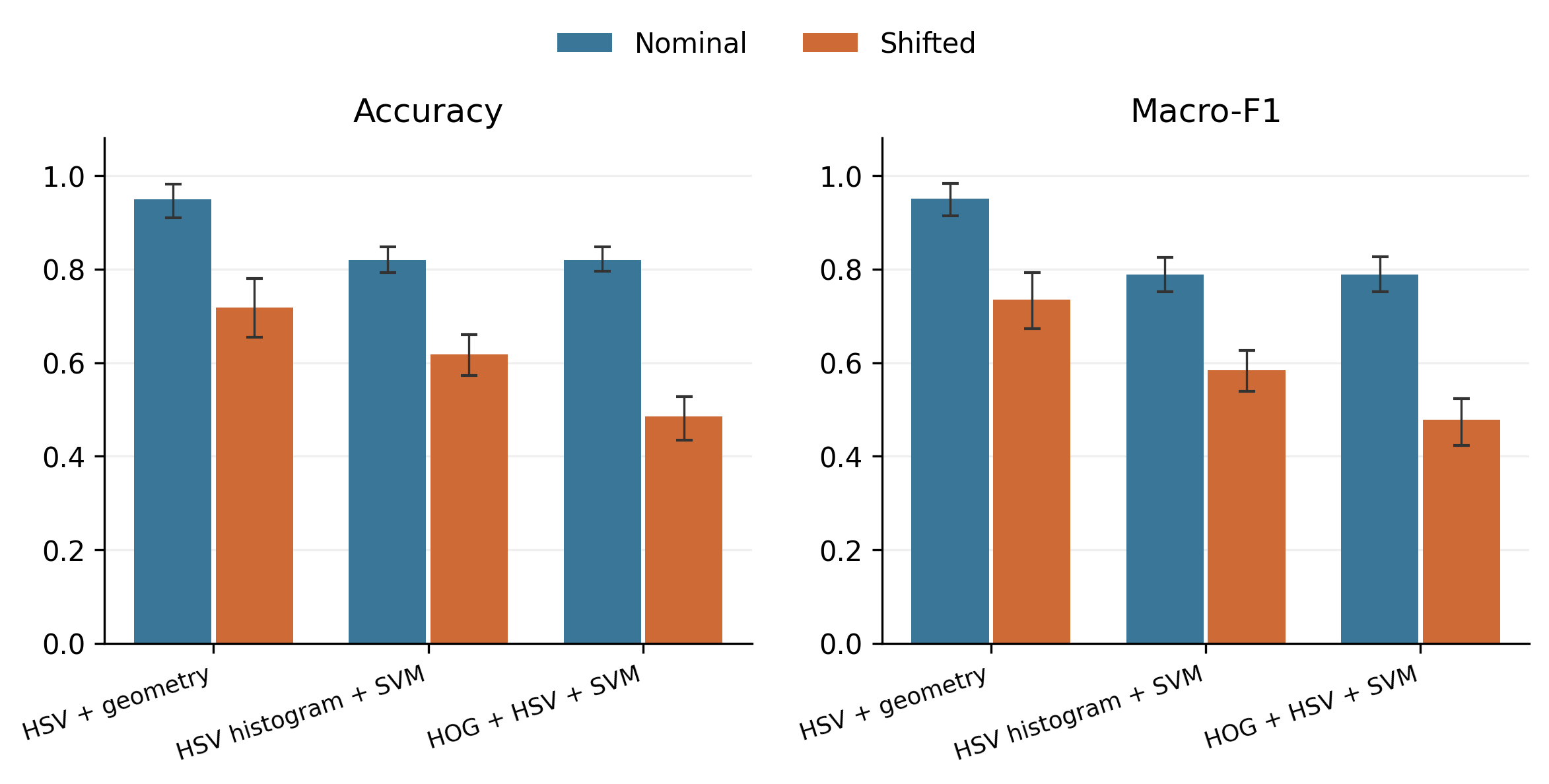}
\caption{Executed classification results under nominal and shifted
synthetic distributions. Error bars use whole-scene bootstrap
resampling. The explicit baseline performs better than both learned
models; the graph is not a field-performance comparison.}
\end{figure}

The HOG-plus-color model artifact occupies 98,188 bytes. Feature
extraction and one-image prediction take a median of 3.466 ms and a 95th
percentile of 10.542 ms across 300 timed calls on a shared x86-64 Xeon
host, after warm-up and with a one-thread BLAS limit. The timing
excludes camera acquisition, image decoding, board localization, and
control. It is not Raspberry Pi 5 timing. A compact artifact and short
host inference time do not compensate for poor negative rejection.

\begin{figure}
\centering
\includegraphics[width=6.45in,height=\textheight]{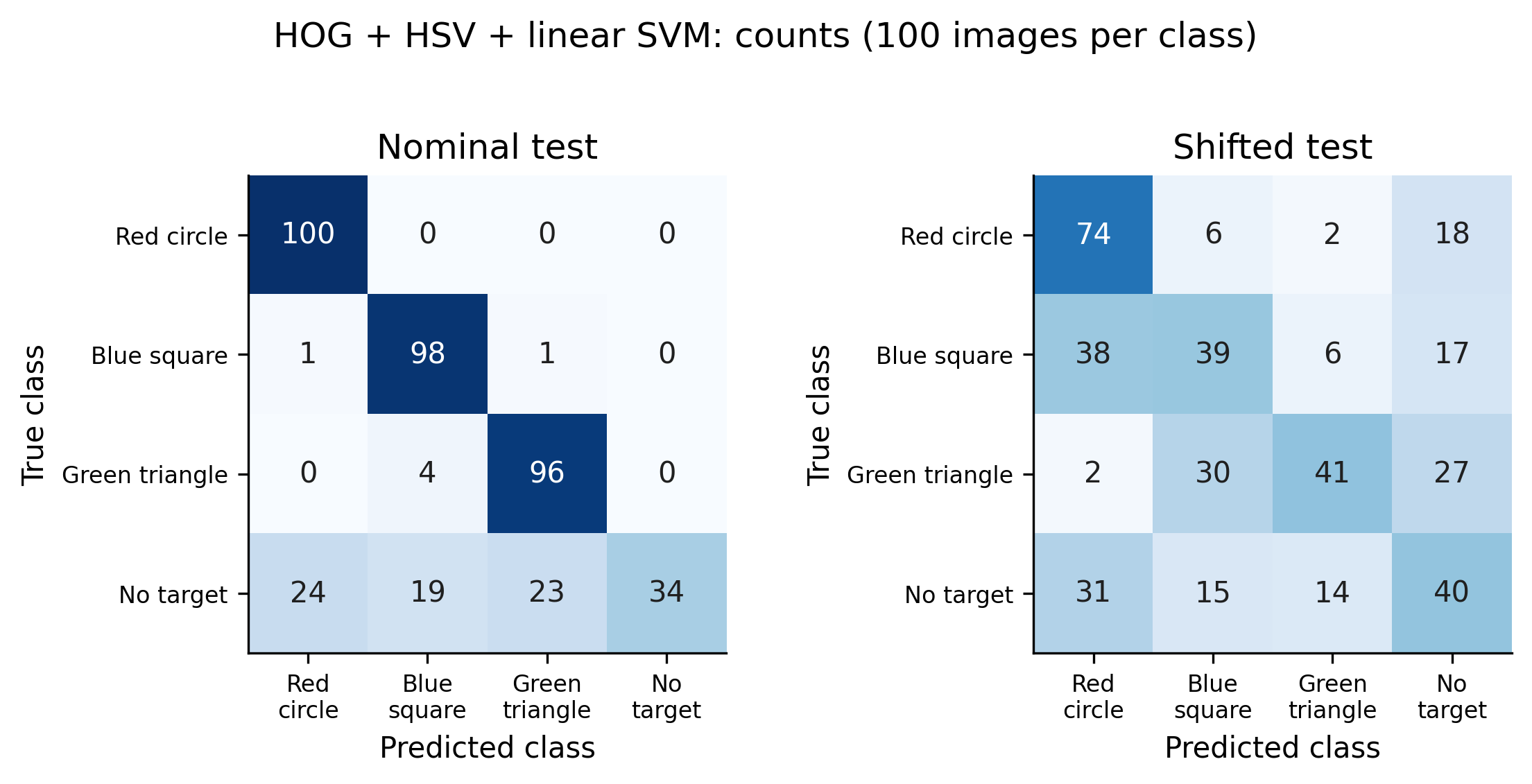}
\caption{Confusion matrices from the executed synthetic benchmark.
Counts correspond to 100 images per true class in each test set. The
no-target column/row behavior is essential when assessing whether
classification could assist board association.}
\end{figure}

\hypertarget{operational-boundary-and-next-experiment}{%
\Needspace{3\baselineskip}
\subsection{9.4. Operational boundary and next
experiment}\label{operational-boundary-and-next-experiment}}

The accompanying adapter consumes a supplied region of interest from an
original unfiltered RGB frame, resizes it, extracts features, and
reports the class and an uncalibrated margin gap. It does not
autonomously locate the board, produce a verified confidence
probability, issue NASA commands, or control the motors. Classification
is therefore a proposed aid to target labeling and later board
association, not a demonstrated dependency of the mandatory command
path.

Fixed target color--shape associations mean that the experiment does not
establish color-independent shape recognition. Mismatched negatives
reduce the shortcut but remain a finite generator-defined mixture. Real
printed markers introduce reflectance, weathering, lens response,
automatic exposure, occlusion, motion, and clutter that the present
generator does not validate. Resizing an ideal supplied crop also avoids
the practical problem of detecting and cropping a small board in a full
frame.

The next useful experiment is a real-camera dataset with separate
acquisition sessions for training and testing, including wrong-color
shapes, blank boards, background clutter, partial views, multiple
ranges, and direct sunlight or shade. Report localization errors
separately from crop-classification errors. If a rejection threshold is
introduced, fit it without using the final test set and report false
acceptance as well as missed targets. Only target-hardware measurements
can establish latency within the command budget. Until those checks are
complete, the geometric baseline is the more defensible comparison point
and ML remains an evaluated research extension.

\hypertarget{integrated-verification-cost-and-research-limitations}{%
\Needspace{4\baselineskip}
\section{10. Integrated Verification, Cost, and Research
Limitations}\label{integrated-verification-cost-and-research-limitations}}

\hypertarget{verification-matrix}{%
\Needspace{3\baselineskip}
\subsection{10.1. Verification matrix}\label{verification-matrix}}

The analytical and software work identifies the measurements that carry
the most decision value. A measured full-charge speed determines limiter
settings. A sustained loaded-motor test addresses the climb-torque
question. Corner weights and suspension travel identify the actual
support geometry. Voltage traces under simultaneous loads address resets
and data loss. A camera--LiDAR board test addresses whether a
valid-looking number belongs to the requested target. These measurements
should be linked to a fixed hardware and software revision so that later
modifications do not inherit evidence from an obsolete configuration.

\FloatBarrier
\begingroup\small
\begin{longtable}[]{@{}
  >{\raggedright\arraybackslash}p{(\columnwidth - 6\tabcolsep) * \real{0.2500}}
  >{\raggedright\arraybackslash}p{(\columnwidth - 6\tabcolsep) * \real{0.2500}}
  >{\raggedright\arraybackslash}p{(\columnwidth - 6\tabcolsep) * \real{0.2500}}
  >{\raggedright\arraybackslash}p{(\columnwidth - 6\tabcolsep) * \real{0.2500}}@{}}
\caption{Integrated verification activities and present
evidence.}\tabularnewline
\toprule\noalign{}
\begin{minipage}[b]{\linewidth}\raggedright
Subsystem
\end{minipage} & \begin{minipage}[b]{\linewidth}\raggedright
Verification activity
\end{minipage} & \begin{minipage}[b]{\linewidth}\raggedright
Record needed
\end{minipage} & \begin{minipage}[b]{\linewidth}\raggedright
Current status
\end{minipage} \\
\midrule\noalign{}
\endfirsthead
\toprule\noalign{}
\begin{minipage}[b]{\linewidth}\raggedright
Subsystem
\end{minipage} & \begin{minipage}[b]{\linewidth}\raggedright
Verification activity
\end{minipage} & \begin{minipage}[b]{\linewidth}\raggedright
Record needed
\end{minipage} & \begin{minipage}[b]{\linewidth}\raggedright
Current status
\end{minipage} \\
\midrule\noalign{}
\endhead
\bottomrule\noalign{}
\endlastfoot
Envelope and mass & Measure fully equipped rover, including antenna &
Dimension sheet and scale record & CAD dimensions and mass estimate
only \\
Propulsion & Full-charge speed, sustained climb, repeated turns,
stop/failsafe & Speed, current, temperature, stop behavior & Analytical
scenarios only \\
Suspension & Loaded corner force--travel and articulation checks &
Motion ratios, end stops, corner weights & CAD concept and sensitivity
model \\
Printed wheels & Coupon, hub, tire/interface and repeated-load tests &
Process settings, forces, displacement, damage & Literature-informed
test plan \\
Task power & Concurrent camera, display and servo operation & Rail
voltage, pack current, temperature & Energy/current scenarios only \\
RF reception & Complete sequence with motors and converters operating &
Received packets, errors, queue timestamps & Architecture and adapter
contract \\
Imaging & Repeated commands, filter replacement, timestamps and storage
& Fresh images and command-linked records & Mocked software checks \\
Ranging & Known distances, board angles, lighting and wrong surfaces &
Raw frames, calibration, reference definition & Geometric analysis and
adapter contract \\
Vision & Session-separated real-camera dataset & Localization, confusion
matrices, rejection metrics & Synthetic experiment completed \\
Integrated excursion & Full configured mission and immediate onboard
review & Versioned event log and results & Not yet performed \\
\end{longtable}
\endgroup

The table avoids assigning an invented acceptance tolerance. Where the
handbook specifies a condition, that condition governs. Where a
system-level tolerance is absent or ambiguous, it must be established
before the test. A pass/fail statement without a defined quantity,
reference, operating condition, and uncertainty is not useful evidence.

\hypertarget{build-budget-and-design-change-control}{%
\Needspace{3\baselineskip}
\subsection{10.2. Build budget and design-change
control}\label{build-budget-and-design-change-control}}

The supplied component budgets list BDT 72,240.00 for chassis and drive
items and BDT 110,854.30 for task items, giving BDT 183,094.30. These
are planning costs, not purchase receipts or a complete lifecycle
estimate. They exclude unrelated travel and event expenses. Different
earlier worksheets used different currency conversions; reporting the
technical component budget in BDT avoids combining inconsistent
exchange-rate bases.

A design change should update the component list, mass estimate, wiring
diagram, analysis inputs, and test record together. For example,
replacing a motor changes speed and current assumptions; changing a
wheel changes speed, torque leverage, clearance, and print validation;
moving a battery changes load distribution and center of mass; replacing
a camera lens changes pixel geometry and calibration. This controlled
relationship is more valuable than adding unsupported precision to one
table.

The estimated mass is the clearest cross-disciplinary tradeoff. Removing
1.847 kg would reach the 20 lb scoring objective, but thinning a printed
rim or shortening a bearing span without load verification could
compromise reliability. The appropriate sequence is to identify
nonstructural mass, quantify the consequence of each change, and then
recheck strength and operation. The present study does not prescribe an
unvalidated lightening pattern.

\hypertarget{what-the-study-establishes}{%
\Needspace{3\baselineskip}
\subsection{10.3. What the study
establishes}\label{what-the-study-establishes}}

The work establishes mathematical consistency for the declared geometry
and scenario parameters, executed behavior for the reference software
with mocks, and reproducible performance for three classifiers on the
declared synthetic crops. It also provides four editable Blender scenes
and six editable Canva architecture diagrams for communicating the
design. Those visual artifacts explain the proposed mechanism and task
flow; they are not multibody-dynamics results, finite-element solutions,
or records of task completion.

The strongest limitations are the absence of native CAD, measured
component performance, physical camera data, calibrated center of mass,
and full-system trials. The analytical model is quasi-static and uses
idealized contact and optical geometry. Printed-material properties are
not characterized. The supplied motor class lacks sufficient continuous
torque/current information for a completed drivetrain selection. The
servo's total usable pan range and the RC distance reference still need
resolution. The synthetic vision study has no simulation-to-real
validation and reports no benefit from the added HOG model.

These limits define a practical research path. A subsequent experimental
paper should add measured methods, uncertainty budgets, repeated trials,
and failure cases rather than rename the current calculations as
experimental results. Literature supplies methods and comparison
context; it cannot fill missing observations for this particular rover.

\hypertarget{reproducibility-and-artifact-provenance}{%
\Needspace{4\baselineskip}
\section{11. Reproducibility and Artifact
Provenance}\label{reproducibility-and-artifact-provenance}}

The accompanying source package contains the canonical manuscript,
IEEE-style bibliography, numerical scripts and CSVs, reference executor,
synthetic generator, trained classifiers, predictions, figure sources,
Blender project, and Canva editing links. Analytical scripts regenerate
six multipanel figure groups and verify unit identities, wheel-dimension
closure, mass conversion, and equivalent force--velocity versus
torque--speed power. The software suite regenerates its mock-adapter
results. The ML verification reloads saved models, reproduces their test
predictions, recomputes metrics, and checks selected generated-image
hashes.

\begin{minipage}{\linewidth}
\begin{Shaded}
\begin{Highlighting}[]
\ExtensionTok{python}\NormalTok{ research/analysis/rover\_analysis.py}
\ExtensionTok{python}\NormalTok{ research/software/check\_executor.py}
\BuiltInTok{cd}\NormalTok{ research/ml}
\ExtensionTok{python} \AttributeTok{{-}m}\NormalTok{ pip install }\AttributeTok{{-}r}\NormalTok{ requirements.txt}
\VariableTok{OPENBLAS\_NUM\_THREADS}\OperatorTok{=}\NormalTok{1 }\VariableTok{OMP\_NUM\_THREADS}\OperatorTok{=}\NormalTok{1 }\DataTypeTok{\textbackslash{}}
  \ExtensionTok{python}\NormalTok{ vision\_experiment.py }\AttributeTok{{-}{-}output}\NormalTok{ reproduced}
\end{Highlighting}
\end{Shaded}
\end{minipage}\par

The supplied CAD screenshots are preserved as design evidence. New
Blender scenes are explicitly procedural approximations from those
views, with embedded parameters and source code. The Blender task scenes
and the synthetic classifier dataset are separate products: no
classifier result is attributed to the task-scene renders. Canva block
diagrams are retained as editable designs; their rendered figure
provenance is recorded with the source package. Published comparison
figures are limited to the two identified CC BY 4.0 reproductions, with
original figure numbers and attribution in their captions. Copyrighted
full-text downloads are not redistributed as part of the package.

The literature log preserves the 20 queries, metadata deduplication,
selected references, access levels, and claim limits. Numerical and
artifact checksums support reproducibility and version identification;
they are not certificates of physical correctness. All calculated,
software-tested, and proposed elements remain distinguishable in the
manuscript and data files.

\FloatBarrier

\hypertarget{conclusion}{%
\Needspace{4\baselineskip}
\section{12. Conclusion}\label{conclusion}}

This study presents a traceable design and software feasibility
assessment of a four-wheel rover with commercial differential drive and
autonomous execution of received imaging and ranging commands. The
latest supplied geometry supports consistent dimensional calculations,
while the estimated 10.919 kg mass exceeds the 20 lb scoring objective
by 1.847 kg. A nominal 200 RPM wheel speed implies 8.105 km/h, showing
that the motor label and an assumed loaded-speed reduction are
insufficient to establish the speed limit. The calculated 20° climb case
requires at least 1.255 N·m per wheel under the declared resistance
assumption, before turning, transient and other unmodeled demands.

The sensing analysis shows that nominal image sampling is only one part
of task feasibility. At 50 ft, a 1° aiming error shifts the ranging ray
by approximately 266 mm, making target association, calibration, and the
distance reference central integration requirements. Separate battery
and regulator branches, correct low-side return routing, bounded command
execution, fresh data, and visible failure records provide a coherent
implementation basis. Thirty mock-adapter checks verify
reference-software behavior without claiming hardware execution.

The synthetic experiment supplies a reproducible negative comparison:
HSV-and-geometry rules outperform both linear SVM models, and adding HOG
reduces shifted accuracy rather than improving it. Learning-assisted
recognition therefore remains an extension requiring real-camera
evaluation. Together, the calculations, code, and visual models identify
a concrete route from the supplied design to a testable prototype.
Physical mobility, electrical, sensing, and complete-sequence
measurements are the remaining evidence needed to establish actual rover
performance.

\FloatBarrier

\end{document}